\documentclass{article}

\usepackage{PRIMEarxiv}

\usepackage[utf8]{inputenc} 
\usepackage[T1]{fontenc}    
\usepackage{hyperref}       
\usepackage{url}            
\usepackage{booktabs}       
\usepackage{amsfonts}       
\usepackage{nicefrac}       
\usepackage{microtype}      
\usepackage{lipsum}
\usepackage{fancyhdr}       
\usepackage{graphicx}       
\graphicspath{{media/}}     
\usepackage{amssymb}
\usepackage{amsmath}
\usepackage{hyperref}
\usepackage{url}
\usepackage{booktabs}
\usepackage{amsfonts}
\usepackage{nicefrac}
\usepackage{microtype}
\usepackage{xcolor}
\usepackage{enumitem}
\usepackage[titles]{tocloft}
\usepackage{graphicx}
\usepackage{subcaption}
\usepackage{amsthm}
\usepackage{mathrsfs}
\usepackage{bm}
\usepackage{siunitx}
\usepackage{multirow}
\usepackage{float}
\usepackage{algorithm}
\usepackage{algpseudocode}
\usepackage{indentfirst}
\usepackage{array}
\usepackage{colortbl}
\definecolor{shadow}{gray}{0.9}
\definecolor{white}{gray}{1.0}
\definecolor{myblue}{rgb}{0.0, 0.2, 0.7}
\newcolumntype{M}[1]{>{\centering\arraybackslash}m{#1}}
\usepackage{tikz}
\usetikzlibrary{shapes.geometric}
\newcommand{\Stars}[2][fill=yellow,draw=orange]{\begin{tikzpicture}[baseline=-0.35em,#1]
\foreach \X in {1,...,3}
{\pgfmathsetmacro{\xfill}{min(1,max(1+#2-\X,0))}
\path (\X*1.1em,0) 
node[star,draw,star point height=0.25em,minimum size=1em,inner sep=0pt,
path picture={\fill (path picture bounding box.south west) 
rectangle  ([xshift=\xfill*1em]path picture bounding box.north west);}]{};
}
\end{tikzpicture}}

\title{\LARGE\textsc{
    \textbf{Adapformer: Adaptive Channel Management for Multivariate Time Series Forecasting}
}}

\author{%
  Yuchen Luo$^1$, Xinyu Li$^2$, Liuhua Peng$^1$, Mingming Gong\thanks{$^*$Corresponding author}$^*$ \\
  $^1$School of Mathematics and Statistics, $^2$School of Computing and Information Systems \\
  The University of Melbourne, Melbourne, Parkville VIC 3052, Australia \\
  \texttt{\{stluo, xl5\}@student.unimelb.edu.au} \\
  \texttt{\{liuhua.peng, mingming.gong\}@unimelb.edu.au} \\
}

\begin{document}
\maketitle

\begin{abstract}
In multivariate time series forecasting (MTSF), accurately modeling the intricate dependencies among multiple variables remains a significant challenge due to the inherent limitations of traditional approaches. Most existing models adopt either \textbf{channel-independent} (CI) or \textbf{channel-dependent} (CD) strategies, each presenting distinct drawbacks. CI methods fail to leverage the potential insights from inter-channel interactions, resulting in models that may not fully exploit the underlying statistical dependencies present in the data. Conversely, CD approaches often incorporate too much extraneous information, risking model overfitting and predictive inefficiency. To address these issues, we introduce the Adaptive Forecasting Transformer (\textbf{Adapformer}), an advanced Transformer-based framework that merges the benefits of CI and CD methodologies through effective channel management. The core of Adapformer lies in its dual-stage encoder-decoder architecture, which includes the \textbf{A}daptive \textbf{C}hannel \textbf{E}nhancer (\textbf{ACE}) for enriching embedding processes and the \textbf{A}daptive \textbf{C}hannel \textbf{F}orecaster (\textbf{ACF}) for refining the predictions. ACE enhances token representations by selectively incorporating essential dependencies, while ACF streamlines the decoding process by focusing on the most relevant covariates, substantially reducing noise and redundancy. Our rigorous testing on diverse datasets shows that Adapformer achieves superior performance over existing models, enhancing both predictive accuracy and computational efficiency, thus making it state-of-the-art in MTSF.
\end{abstract}


\section{Introduction}
\label{introduction}

Multivariate Time Series Forecasting (MTSF) is a pivotal branch of statistical learning and data mining focused on predicting future values of multiple interrelated variables based on historical observations. This field plays a crucial role across various domains such as economics \cite{granger2014forecasting}, finance \cite{tsay2005analys}, meteorology, and engineering \cite{wiener1949extrapolation}, where understanding spatial-temporal dynamics is essential. With the recent advancements in deep learning where sophisticated neural architectures are introduced, Transformer architecture \cite{vaswani2017attention} has emerged as particularly influential. Its strong capability in effectively retrieving sequential dependencies by leveraging self-attention mechanisms has set new benchmarks for accuracy and scalability in MTSF \cite{vaswani2017attention, zhang2023crossformer, liu2023itransformer}. Existing approaches treat observations of multiple variables at each time step as separate tokens~\cite{wu2021autoformer, zhou2022fedformer, serra2018towards} and propose innovative Transformer-based architectures to discern point-wise temporal dependencies. However, multiple measurements in real-world scenarios captured over the same period frequently represent misaligned physical events. As a result, these methodologies often face challenges in accurately modeling the intricate relationships and dynamics among variables, potentially compromising the fidelity of their analytical outcomes. In response, researchers have raised the primary challenge of simultaneously modeling both intra-series (temporal) dependencies and inter-series (cross-variable) relationships among numerous covariates \cite{wang2024card, zhang2023crossformer, wu2022timesnet, zeng2023transformers}.

\begin{figure}
    \centering
    \includegraphics[width=1\linewidth]{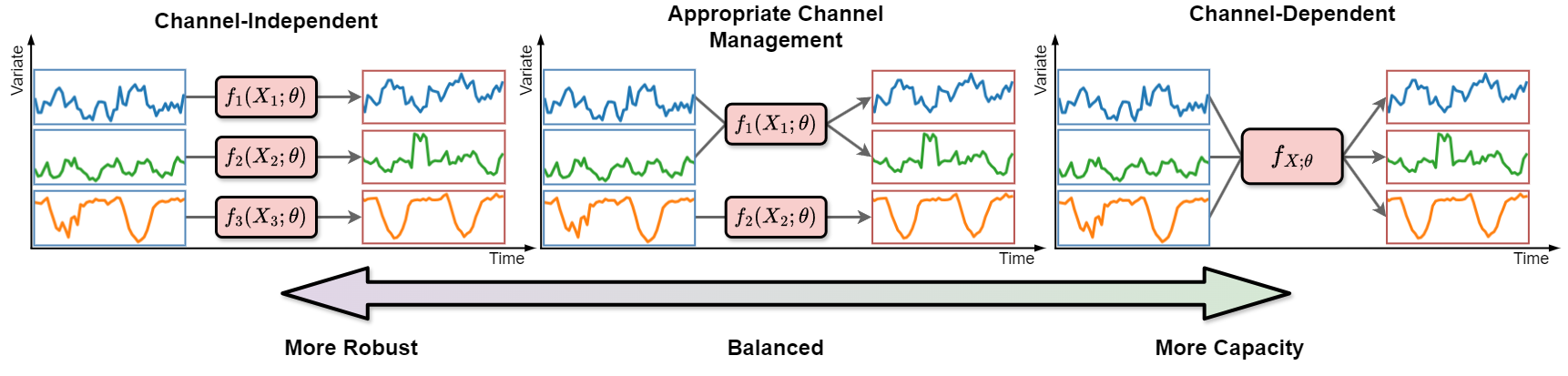}
    \caption[Channel Independent \& Dependent Strategy]{Channel Independent Strategy with more Robustness (Left) and Channel Dependent Strategy with more Model Capacity (Right). An appropriate channel management should balance in the middle.}
    \label{cicd}
\end{figure}

Furthermore, existing methodologies in multivariate time series forecasting can be broadly classified into channel-independent (CI) and channel-dependent (CD) approaches. As illustrated in Figure~\ref{cicd}, CI methods\cite{nie2022time, zeng2023transformers} treat each variable as an isolated target, focusing on capturing temporal patterns within individual channels. While this enhances channel-specific learning, it often overlooks inter-channel interactions~\cite{han2024capacity}, thereby limiting the model's ability to generalize across variables. Conversely, CD methods~\cite{wang2024timemixer, zhang2023crossformer, wang2024card} aim to integrate information across all covariates to effectively model inter-series dependencies. However, empirical research~\cite{nie2024channel} has shown that incorporating an excessive number of variables can introduce substantial noise, adversely affecting the model's performance. Consequently, purely channel-independent methods may lack model capacity while purely channel-dependent methods may lose robustness. Therefore, it is imperative to implement selective channel management strategies to balance the utilization of informative signals while mitigating the impact of noise from the increasing number of variables. 

Moreover, prior studies~\cite{zhang2023crossformer, wang2024card, wang2024timemixer} have extensively focused on developing novel encoder~\cite{chen2018encoder} modules within the Transformer architecture, incorporating advanced attention mechanisms that are better suited to the task of multivariate time series forecasting. For instance, traditional point-wise attention~\cite{vaswani2017attention}, which targets singular timestamps, often fails to capture extended temporal dynamics due to the limited insights available at any single point in real-world observations. In response, patch-wise attention~\cite{nie2022time} broadens the scope by covering wider temporal intervals, thus significantly enhancing the model's capability to comprehend and incorporate extensive temporal semantics. However, there has been comparatively less emphasis on the embedding and decoding stages of the Transformer architecture. From the perspective of channel management, the roles of the embedding and decoding stages are also crucial. Specifically, optimizing the embedding layer is pivotal as it generates tokenized representations that integrate key information across various channels. This foundational step ensures that the subsequent encoding processes are equipped with a rich, contextually nuanced input leading to larger model capacity. Meanwhile, enhancements in the decoding stage play a complementary role by precisely filtering out noise that can arise from excessive channel inputs, thus refining the model's output. These adjustments ensure that the final predictions are not only cleaner, but also more aligned with the inter-series dynamics identified by the encoder, leading to more reliable and robust forecasting results.

In response to the aforementioned challenges and motivations,  this study innovatively synthesizes CI and CD methodologies to harness their collective strengths while mitigating their respective limitations. We introduce a sophisticated model, the Adaptive Forecasting Transformer (\textbf{Adapformer}), which strategically enhances information fusion for highly correlated variables and effectively suppresses less relevant data streams, thereby optimizing the interplay between CI and CD modalities. Our approach is characterized by a novel dual-stage architecture within the conventional Transformer framework, incorporating two pivotal modules: the \textbf{A}daptive \textbf{C}hannel \textbf{E}nhancer (\textbf{ACE}) and the \textbf{A}daptive \textbf{C}hannel \textbf{F}orecaster (\textbf{ACF}). The initial stage involves embedding along individual variables, thereby allowing the attention mechanism to meticulously capture and analyze inter-channel dependencies. Subsequently, the ACE module is designed to advance this setup by prioritizing channel-specific learning, reflecting the essential temporal dependencies through a learnable low-rank approximation~\cite{hu2021lora}. This selective integration process within each token not only aims to enrich its representational efficacy, but also unveils latent temporal structures across variables. Transitioning to the decoding phase, our ACF module deviates from traditional all-to-all prediction strategies by adopting a discerning approach that selectively engages the most salient covariates for each channel prediction. This targeted utilization strategy is conceptualized not merely as a novel decoding mechanism but also as an innovative predictive paradigm that can be seamlessly integrated into existing models to elevate their performance. Experimentally, our Adapformer achieved state-of-the-art performance on 7 real-world benchmark datasets widely used in MTSF. The contribution of this study can be summarized in three folds:

\begin{itemize}[itemsep=0pt, topsep=1pt]
    \item We present the Adapformer, an innovative Transformer-based model that seamlessly combines the strengths of channel-independent and channel-dependent strategies in multivariate time series forecasting.
    \item We design a dual-stage architecture featuring the Adaptive Channel Enhancer (ACE) and the Adaptive Channel Forecaster (ACF), which enhance information utilization and can be seamlessly integrated into other MTSF models.
    \item We demonstrate the effectiveness of exploring a new prediction paradigm in multivariate time series forecasting, encouraging future research to build upon this direction.
\end{itemize}

\section{Related Work}
\label{related_work}

Advancements in multivariate time series forecasting (MTSF) have spurred the development of innovative methods designed to more effectively model complex relationships. Moving beyond traditional approaches, researchers have introduced novel techniques that enhance both model performance and computational efficiency. One line of research focuses on patch-based methods. PatchTST \cite{nie2022time} addresses the limitations of point-wise temporal tokens by segmenting time series data into patches, each containing multiple consecutive time steps. By embedding these patches into tokens and applying attention mechanisms, the model captures local temporal patterns while reducing sequence length and computational complexity. However, its channel-independent nature limits the exploration of complex inter-series dependencies and adaptability to different time scales. To overcome these limitations, Crossformer \cite{zhang2023crossformer} extends the patch-wise approach by integrating cross-attention mechanisms. It captures interactions between different variables across patches, modeling rich semantics along both temporal and cross-variable dimensions. While this enhances the model's capacity, it also introduces additional computational complexity and potential overfitting due to the larger scale of information utilization. Alternatively, the iTransformer \cite{liu2023itransformer} employs a series-wise attention strategy by embedding each variate across the entire time horizon into a unified token. This inverted perspective effectively captures global temporal trends and addresses time misalignment issues where variables may exhibit asynchronous behaviors. Notably, it maintains the simplicity of the vanilla Transformer architecture without introducing additional modifications, solely altering the modeling dimension.

Beyond attention-based mechanisms, channel-based innovations have significantly contributed to MTSF. DLinear \cite{zeng2023transformers} challenges the necessity of complex architectures by using simple multilayer perceptrons (MLPs) to predict each variable independently. By modeling each channel separately and decomposing inputs into trend and seasonality components, DLinear outperforms more sophisticated Transformer variants, highlighting the effectiveness of channel-independent (CI) approaches. Models like TimesNet and TimeMixer aim to enhance channel-specific exploration within the channel-dependent (CD) framework. TimesNet~\cite{wu2022timesnet} introduces the TimesBlock, leveraging convolutional neural networks (CNNs) to capture variations within and between multiple periods by transforming time series into a two-dimensional representation through multi-periodicity decomposition. TimeMixer \cite{wang2024timemixer} employs a multiscale decomposition approach with MLPs, stacking down-sampled data into layers and applying mixing strategies to balance short-term and long-term dependencies across channels. Collectively, these models strive to balance the strengths of CI and CD approaches. By harmonizing intra-series and inter-series dynamics, they enhance forecasting performance and highlight opportunities for future research to further optimize the interplay between CI and CD strategies. This categorization is summarized in Table~\ref{table_cicd_summary}, which compares representative models along their CI and CD strengths, architectural complexity, and design motivations. This progress underscores the importance of innovative architectural designs in advancing both model efficiency and accuracy in multivariate time series forecasting.

Unlike prior approaches, our proposed Adapformer retains the native Transformer components unchanged. Instead, we enhance performance through strategic manipulation of token embeddings and selective channel management, offering a novel balance between channel-independent and channel-dependent methods without adding complexity to the core architecture. 

\begin{table}[H]
    
    \centering
    \caption{Summary of representative multivariate time series forecasting models characterized by their emphasis on channel-independent (CI) and channel-dependent (CD) modeling strategies. CI and CD strengths reflect each model’s design for intra- and inter-variable dependency learning. Model complexity refers to training-time cost, with “high” indicating greater memory and runtime requirements. Adapformer (Ours) is included to contextualize our method under the CI/CD categorization, illustrating its balanced dual-stage design tailored to both dependency types.}
    \label{table_cicd_summary}
    \resizebox{\textwidth}{!}
    {
    \small
    \begin{tabular}{l|c|c|c|l}
        \toprule
        Model & CI Strength & CD Strength & Complexity & Notes \\
        \midrule
        Adapformer (Ours) & \Stars[]{3} & \Stars[]{3} & Low-Medium & Dual-stage: ACE for selective CI, ACF for refined CD signal utilization.\\
        \cmidrule{1-5}
        TimeMixer (2024)~\cite{wang2024timemixer} & \Stars[]{3} & \Stars[]{1} & Medium & Multi-scale MLP with moderate CI and strong CD via mixed predictors.\\
        \cmidrule{1-5}
        iTransformer (2024)~\cite{liu2023itransformer} & \Stars[]{0.5} & \Stars[]{3} & Low-Medium & Series-level embedding enables CD learning over global dependency.\\
        \cmidrule{1-5}
        PatchTST (2023)~\cite{nie2022time} & \Stars[]{3} & \Stars[]{0} & Medium & Patch-wise segmentation allows localized CI patterns with minor CD effect.\\
        \cmidrule{1-5}
        Crossformer (2023)~\cite{zhang2023crossformer} & \Stars[]{2} & \Stars[]{2} & High & Strong CD through cross-attention, enhancing inter-series reasoning. \\
        \cmidrule{1-5}
        CARD (2023)~\cite{wang2024card} & \Stars[]{2} & \Stars[]{2} & High & Dual-path MLPs jointly extract intra- and inter-channel signals.\\
        \cmidrule{1-5}
        TimesNet (2023)~\cite{wu2022timesnet} & \Stars[]{2} & \Stars[]{0.5} & High & Multi-periodicity modeling targeting intra-\&inter-channel interactions. \\
        \cmidrule{1-5}
        DLinear (2023)~\cite{zeng2023transformers} & \Stars[]{2} & \Stars[]{0} & Low & Pure CI via separate MLPs per channel without fusion.\\
        \cmidrule{1-5}
        FEDformer (2023)~\cite{zhou2022fedformer} & \Stars[]{1.5} & \Stars[]{0.5} & Medium & Frequency-based CD with Fourier attention across series.\\
        \cmidrule{1-5}
        Autoformer (2022)~\cite{wu2021autoformer} & \Stars[]{1.5} & \Stars[]{0.5} & Low-Medium & Decomposes CI trends jointly across all channels for CD.\\
        \cmidrule{1-5}
        Transformer (2021)~\cite{vaswani2017attention} & \Stars[]{1} & \Stars[]{0.5} & Low & Baseline attention model without channel-specific design.\\
        \bottomrule
        
    \end{tabular}
    }
\end{table}

\section{Adapformer}
\label{adapformer}

\paragraph{\textbf{Problem Definition}} We begin by formally defining our central task: long-term multivariate time series forecasting (MTSF). This involves using historical observations of multiple interrelated variables to predict their future values concurrently over a forecasting horizon that typically exceeds the length of the input sequence. Compared to standard or short-term forecasting, long-term MTSF adds complexity due to the necessity of capturing both long-range temporal dependencies and inter-variable relationships over extended periods. Modeling these dependencies while maintaining predictive accuracy introduces significant challenges, requiring advanced techniques to effectively manage the inherent interactions and dynamics in long-term multivariate forecasting tasks. Consider an input multivariate time series with $N$ distinct variates, represented as a matrix as $\mathbf{X}=\{\mathbf{x}_1, \mathbf{x}_2,\ldots, \mathbf{x}_T\} \in \mathbb{R}^{T \times N}$, where the lookback window size $T$ stands for the length of observations and each $\mathbf{x}_t \in \mathbb{R}^N$ corresponds to $N$ concurrent variates observed at time $t$. Also, consider the subsequent time series of $\mathbf{X}$ over a future horizon $L$, denoted as $\mathbf{Y}=\{\mathbf{x}_{T+1}, \mathbf{x}_{T+2}, \ldots, \mathbf{x}_{T+L}\} \in \mathbb{R}^{L \times N}$. Given a forecasting model $f_\theta$ parameterized by $\theta$ that maps the historical time series $\mathbf{X}$ to its $L$ steps future forecast as $\mathbf{\hat{Y}} = f_\theta(\mathbf{X})$, where $\mathbf{\hat{Y}}=\{\mathbf{\hat{x}}_{T+1}, \mathbf{\hat{x}}_{T+2}, \ldots, \mathbf{\hat{x}}_{T+L}\} \in \mathbb{R}^{L \times N}$, the objective of time series forecasting is to optimize $\theta$ such that the forecast $\mathbf{\hat{Y}}$ closely approximates the true future value $\mathbf{Y}$.

\paragraph{\textbf{Structure Overview}}

\begin{figure}
    \centering
    \includegraphics[width=1\linewidth]{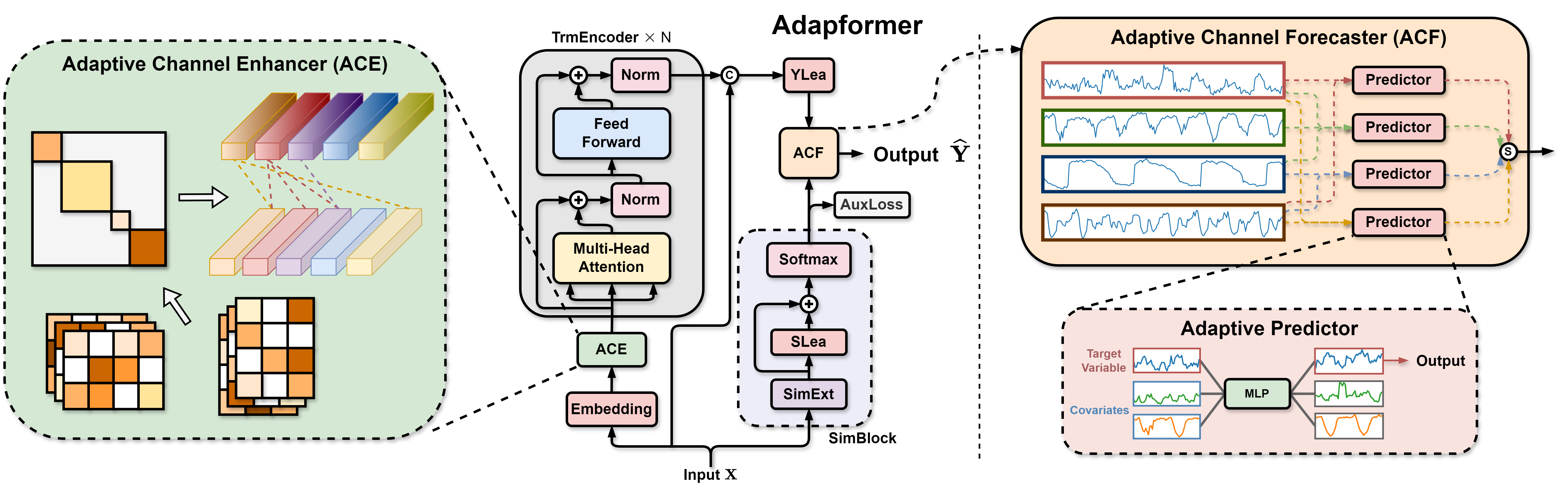}
    \caption{An overview of the proposed Adapformer architecture. The raw inputs are first embedded and subsequently refined by the Adaptive Channel Enhancer (ACE), which enriches each token’s representational capacity. Canonical Transformer encoders then captures dependencies among these enhanced tokens, producing encoded representations that are passed to the decoder - Adaptive Channel Forecaster (ACF) - to predict future time-series outputs. Meanwhile, a separate similarity block (SimBlock) processes the raw inputs to explicitly model inter-sequence relationships for future predictions, providing an auxiliary output used both for additional training supervision and to further guide the ACF’s forecasting process. the ACE module.}
    \label{fig14}
\end{figure}

Our proposed \textbf{Adapformer} introduces an innovative solution to MTSF that enhances the traditional Transformer architecture through strategic manipulation of token embeddings and selective channel management. We commence our modeling process by embedding each channel separately following previous study\cite{liu2023itransformer}, allowing the model to maintain distinct channel identities essential for capturing nuanced inter-channel dependencies. To effectively model both spatial and temporal dimensions, we integrate these aspects within each token through the Adaptive Channel Enhancer (ACE) right after the embedding layer. These enriched tokens are then processed by the canonical Transformer encoder stack, where we implement instance normalization to preserve the unique statistical characteristics of each variable. This modification follows the discussion in previous research. In the context of MTSF, each variable represents different physical measurements, and using layer normalization \cite{ba2016layer} risks blending their unique statistical characteristics. To preserve this uniqueness, we employ Instance Normalization \cite{ulyanov2017instance} in the encoder, normalizing each time series independently. This step helps retain variate-specific properties and enhances the model's ability to handle non-stationary time series effectively. In the prediction phase, we address the challenge of high-dimensional data in MTSF by employing an adaptive channel constraint in our decoding strategy. Unlike previous methods \cite{nie2022time, liu2023itransformer} that either predict in isolation per channel or utilize all channels indiscriminately, our Adaptive Channel Forecaster (ACF) selectively uses the most relevant channels for forecasting each specific target channel. This approach significantly cuts down on redundant information and noise, enhancing model robustness and reducing the likelihood of overfitting. To determine the level of inter-series relevance, we introduce a parallel module called SimBlock, designed to extract future inter-series correlations from the raw input data. Figure \ref{fig14} illustrates the seamless integration of our key modules: the Adaptive Channel Enhancer (ACE), the SimBlock, and the Adaptive Channel Forecaster (ACF). This holistic design effectively manages channel contributions throughout the modeling process, enabling our system to maintain robustness while capturing a richer spectrum of multivariate interactions. Consequently, our model represents a significant advancement in the field of MTSF. \noindent To predict the future series $\mathbf{Y} \in \mathbb{R}^{L \times N}$ of length $L$ with a given time series $\mathbf{X} \in \mathbb{R}^{T \times N}$ with look back window size $T$ and number of variates $N$, the above architecture can be formulated as follows:

\begin{equation}
    \begin{split}
        &\mathbf{X}_{\operatorname{Norm}} = \operatorname{RevIN}(\mathbf{X}), \\
        &\mathbf{X}_{\operatorname{Emb}} = \operatorname{Embedding}(\mathbf{X}_{\operatorname{Norm}}^\top), \\
        &\mathbf{X}_{\operatorname{enc}}^0 = \operatorname{ACE}(\mathbf{X}_{\operatorname{Emb}}), \\
        &\mathbf{X}_{\operatorname{enc}}^i = \operatorname{TrmEncoder}_i(\mathbf{X}_{\operatorname{enc}}^{(i-1)}), i = 1, 2, \ldots, J, \\
        &\mathbf{\hat{Y}}^\top = \operatorname{ACF}(\mathbf{X}_{\operatorname{enc}}^J, \operatorname{SimBlock}(\mathbf{X}_{\operatorname{Norm}})), 
    \end{split}
\end{equation} where the reversible instance normalization~\cite{RevIn} follows the standard practice established in previous works \cite{liu2022non}: $\operatorname{RevIN}(\mathbf{X}) = \frac{\mathbf{X} - E[\mathbf{X}]}{\operatorname{Std}(\mathbf{X})}$. The functions $\operatorname{Embedding}: \mathbb{R}^T \mapsto \mathbb{R}^D \text{ and } \operatorname{ACE}: \mathbb{R}^D \mapsto \mathbb{R}^D$ work in tandem to construct the desired tokens in the hidden dimension $D$. Specifically, the $i$-th Transformer encoder is denoted as $\operatorname{TrmEncoder}_i: \mathbb{R}^{N \times D} \mapsto \mathbb{R}^{N \times D}$. Parallel to this process, the raw input time series is fed into the $\operatorname{SimBlock}: \mathbb{R}^{T\times N} \mapsto \mathbb{R}^{N\times N}$ which predicts future channel correlations. These predicted correlations are leveraged by the $\operatorname{ACF}: \mathbb{R}^{N \times D} \mapsto \mathbb{R}^{N \times L}$ to generate the final predictions. The full architecture and training logic are outlined in the pseudocode provided in \ref{appendix_pseudocode}.

\paragraph{\textbf{Adaptive Channel Enhancer}}

Targeting the motivation that we previously discussed, our primary objective at this stage is to facilitate the integration over both intra-series (temporal) and inter-series (spatial) dimensions for the embedded representation. Based on the crucial ability in capturing token-wise relationships of the self attention mechanism, we explore two principle methodologies to achieve the enriched tokens:
\begin{enumerate}
    \item \textbf{Channel-wise Embedding}: First embed each channel independently then emphasis the channel-specific patterns within each token.
    \item \textbf{Time-step Embedding}: Embed all observations from the same time step before adding in cross-variate interactions.
\end{enumerate}
Based on the discussions in the preceding sections, we believe any cross-variate embedding without adequate constraint is prone to overwhelm the system with latent noise. Meanwhile, in view of the success of the inverse-embedding approach introduced by Liu et al. \cite{liu2023itransformer}, we follow this emphasize of preserving the unique statistical characteristics of each variate. Therefore we adopt the former design among the two approaches. Concretely, the initial separation of channel-wise embedding allows the tokens to maintain distinct channel identities early on \cite{nie2022time, nie2024channel}, which ensures the encoder to retrieve the inter-series dependencies. Therefore the Adaptive Channel Enhancer is intentionally designed to strengthen the channel-specific patterns within each token. Alternatively, it aims to capture the specific underlying temporal dependencies that are crucial for understanding dynamics over time within every single series. However, simply applying a standard channel-wise embedding is often insufficient for time series modeling, as it primarily captures static feature mappings without explicitly enhancing temporal dynamics. In time series data, crucial patterns such as trends, seasonality, and sudden regime shifts are easily overshadowed by noise, especially in high-dimensional multivariate settings. Therefore we employ a low-rank approximation \cite{liu2012robust, davenport2016overview} technique to selectively reinforce token embeddings within each single variate. Low-rank approximation is rooted in linear algebra and matrix factorization techniques, used to reduce the dimensionality of data by approximating a high-dimensional matrix with the product of two lower-dimensional matrices. This technique captures the most significant patterns within the data while discarding noise and less critical information, thereby enhancing the model's ability to focus on essential inter-variable relationships. Given a high-dimensional target matrix $\mathbf{T} \in \mathbb{R}^{m \times n}$, it can be approximated by the product of two lower-dimensional matrices $\mathbf{L} \in \mathbb{R}^{m \times k}$ and $\mathbf{R} \in \mathbb{R}^{k \times n}$, such that $\mathbf{T} \approx \mathbf{L}\mathbf{R}$ with $k \ll \operatorname{min}(m, n)$. The objective is to retain the most salient features of $\mathbf{T}$ while significantly reducing computational complexity. This approximation is particularly effective in scenarios where the data exhibits underlying low-dimensional structures, enabling the model to capture essential patterns without being encumbered by high-dimensional noise. The succeed of such strategy has also been validated and applied in Large Language Models \cite{hu2021lora}. In our context, employing low-rank approximation specifically aids in identifying and emphasizing crucial temporal signals like seasonality and trends within time series data. By reducing the data dimensionality, this technique isolates the core patterns that are most informative for predictions, such as annual cycles in weather or market trends in finance, allowing the model to focus on these significant features while ignoring less relevant noise. More concretely, the rank $r$ in the process intuitively represents the number of independent temporal patterns that the model is allowed to extract and inject into each channel's embedding. A smaller $r$ constrains the model to capture only the most dominant dynamics, such as major trends and seasonal cycles, while naturally filtering out minor fluctuations and noise. Conceptually, $r$ controls the expressiveness of the enhancement: it defines how many principal modes of variation are emphasized in the enriched token representation. Consequently, upon entering the encoder, these enriched tokens are laden with sufficient spatial-temporal patterns, enabling the standard point-wise self attention to effectively discern and capture critical patterns within the data. This ensures that the model can leverage the full context provided by the enhanced embeddings without additional modifications to the attention mechanism. Furthermore, It is also worth noting that although ACE primarily focuses on enhancing intra-channel temporal patterns, the use of shared low-rank projection weights across channels implicitly allows the model to extract limited cross-channel dynamics when beneficial, without the need for dense full-channel interaction.

\paragraph{\textbf{Adaptive Channel Forecaster}}

\begin{figure}
    \centering
    \includegraphics[width=0.8\linewidth]{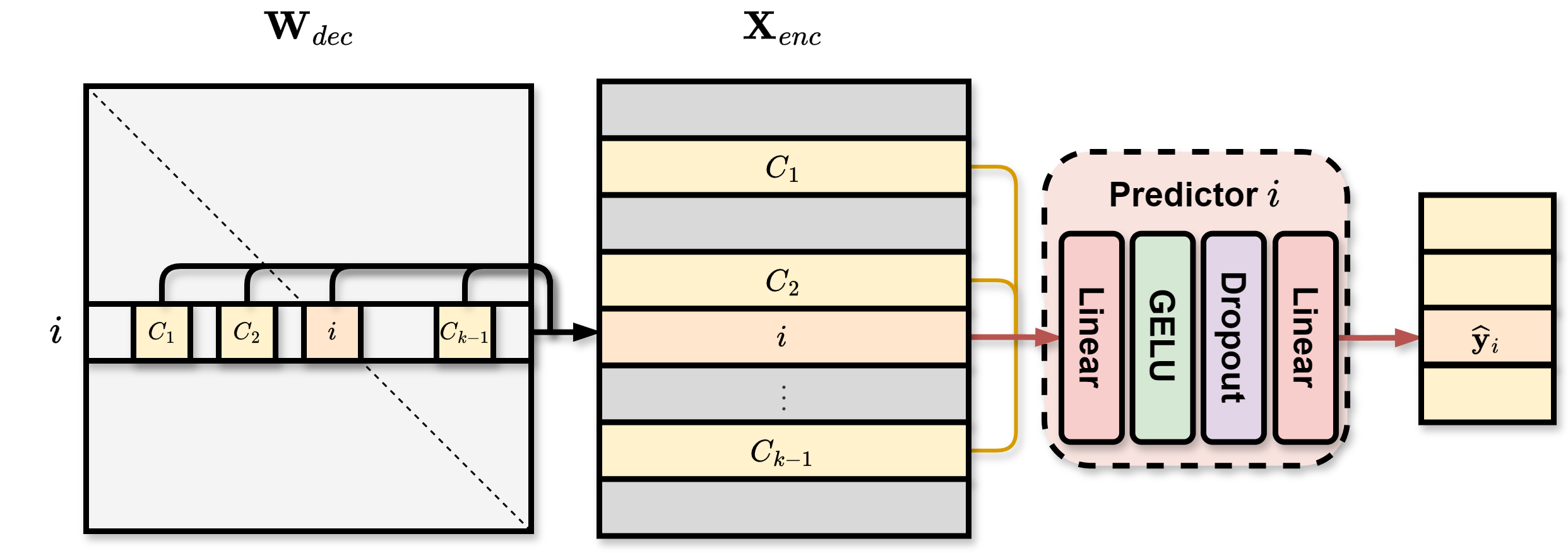}
    \caption{Illustration of the Adaptive Channel Forecaster (ACF). To predict the $i$-th target variable, the model first selects $k-1$ covariates most correlated with the target among future variables according to the results from SimBlock, forming a set of $k$ inputs (including the $i$-th variable itself). These inputs are then processed through a simple MLP with skip connection to produce future sequences, from which only the $i$-th target variable is retained as the final output.}
    \label{fig18}
\end{figure}

In response to the motivation that not all covariates within a dataset exert a positive influence on the prediction of a specific target variable; in some instances, they may even introduce detrimental interference. Nevertheless, most existing methodologies uniformly generate future series for all channels concurrently, thereby engendering mutual noise across extensive variables. More recent research \cite{channelclustering} have also attempted performing clustering and adopting cluster-aware prediction. However, as the encoded representation are often high-dimensional, clustering could not guarantee an ideal outcome. To bridge the gap, the Adaptive Channel Forecaster (ACF) is specially designed for the predictive stage of the Adapformer, aiming to mitigate the excessive noise. 

We employ an innovative element-wise approach that involves sequential generation of future forecasts for each individual channel. Thereby allows an adaptive control over the dynamic selection of the most effective factors pertinent to each specific prediction task. More specifically, for each target variate, we identify the top $k$ most relevant channels based on their similarity relationships which is learned by the SimBlock. This subset of channels, including the target itself, is then input into an independent linear predictor. The linear predictor generates forecasts for all $k$ channels; however, only the prediction corresponding to the target variate is retained from the output. By systematically applying this procedure to each individual variate, we construct the final prediction series by aggregating the preserved target-specific predictions across all channels. Consequently, the ACF ensures that each variate is predicted using these carefully selected subset of covariates, optimizing the balance between comprehensive information utilization and noise mitigation within the forecasting framework. Conceptually, $k$ controls the breadth of contextual information accessible to each target variable: a smaller $k$ enforces a highly selective focus on the most relevant covariates, enhancing robustness against noise, while a larger $k$ allows broader contextual integration at the risk of introducing less pertinent signals. This tunable mechanism enables ACF to dynamically balance information richness and noise suppression according to the task and data complexity. Mathematically, when targeting the $i$-th variable $\mathbf{x}_i$, we adopt the following approach as displayed in Figure \ref{fig18}:
\begin{equation}
    \begin{split}
        \mathcal{C}_i &= \text{TopK}\left(\mathbf{W}_{\operatorname{dec}}[i, :], k\right), \quad \forall i \in \{1, 2, \ldots, N\}. \\
        &= [i, C_1, C_2, ..., C_{k-1}],  \\
        \mathbf{X}_{\mathcal{C}_i} &= \mathbf{X}[\mathcal{C}_i, :], \\
        \mathbf{\hat{Y}}_i &= \operatorname{Predictor}_i(\mathbf{X}_{\mathcal{C}_i}),\\
        \hat{\mathbf{y}}_i &= \mathbf{\hat{Y}}_i[0, :],        
    \end{split}
\end{equation}
where $\mathcal{C}_i \in \mathbb{Z}^{k}$ denotes the indices of the top $k$ selected channels and $\mathbf{X}_{\mathcal{C}_i} \in \mathbb{R}^{k \times D}$ represents the corresponding data subset. The target variate is positioned in the first row. Subsequently, $\mathbf{\hat{Y}}_i$ is the output of the $i$-th predictor, from which only the first channel, $\hat{\mathbf{y}}_i \in \mathbb{R}^{1 \times L}$, corresponding to the target, is retained, where $L$ is the prediction length. Finally, all target predictions are aggregated to form $\mathbf{\hat{Y}} \in \mathbb{R}^{N \times L}$ as the future series:

\begin{equation}
    \mathbf{\hat{Y}} = \begin{bmatrix}
        \hat{\mathbf{y}}_1 \\
        \hat{\mathbf{y}}_2 \\
        \vdots \\
        \hat{\mathbf{y}}_N \\
        \end{bmatrix}.
\end{equation}

\paragraph{\textbf{SimBlock}}

The Similarity Block (SimBlock) constitutes a foundational component of our decoder architecture, meticulously engineered to capture and utilize the intrinsic correlations among multiple variates. The primary motivation behind SimBlock extends beyond the explicit modeling of cross-variate correlations; we posit that each variable, representing distinct physical observations, tends to maintain relatively stable correlation relationships along the temporal dimension. In other words, the inter-dependencies and associations among variables exhibit temporal consistency, exhibiting minimal fluctuations over time. This inherent stability in correlations provides a valuable opportunity to preemptively extract and utilize these relationships to enhance future sequence predictions. Leveraging this insight, SimBlock is designed to explicitly extract the multivariate correlation matrix from the raw input data, thereby capturing the enduring future inter-variable dependencies that are pivotal for accurate forecasting. To operationalize this strategy, SimBlock processes the raw input time series data immediately following normalization and prior to embedding. Given an input matrix $\mathbf{X} \in \mathbb{R}^{T \times N}$ where $N$ represents the number of channels (variates) and $T$ the sequence length, we extract the correlation matrix $\mathbf{W}$:

\begin{equation}
    \begin{split}
        \mathbf{W} = \langle \mathbf{X}^T, \mathbf{X} \rangle \in \mathbb{R}^{N \times N}.
    \end{split}
\end{equation}
This symmetric matrix $\mathbf{W}$ encapsulates the pairwise correlations between variates, effectively quantifying the inter-variable dependencies inherent in the input data. Aiming at estimating the actual true latent relationships embedded in the future series, $\mathbf{W}$ is subjected to a non-linear transformation with skip connection. Subsequently the output goes through a softmax normalization to enforce a scoring structure, quantifying the similarity between pair-wise variables. This normalization step guarantees that $\mathbf{W}_{\operatorname{dec}}$ retains its interpretative role as a weight matrix, emphasizing the most significant correlations and attenuating weaker ones:
\begin{equation}
    \begin{split}
        \mathbf{W}_{\operatorname{dec}} = \operatorname{Softmax}(\mathbf{W} + \operatorname{ReLU}(\operatorname{Linear}(\mathbf{W}))).
    \end{split}
\end{equation}
This normalized weight matrix $\mathbf{W}_{\operatorname{dec}}$ is then forwarded to the subsequent Adaptive Channel Forecaster module to assist in the prediction process. Additionally, to enforce the uniqueness and accuracy of the learned correlations, $\mathbf{W}_{\operatorname{dec}}$ is outputted alongside the actual future sequences and subject to an auxiliary loss function:
\begin{equation}
\label{aux_loss}
    \begin{aligned}
        \mathcal{L}_{\operatorname{aux}} &= ||\mathbf{W}_{y}-\mathbf{W}_{\operatorname{dec}}||^2_2 \\
        &= ||\langle \mathbf{Y}, \mathbf{Y}^T \rangle-\mathbf{W}_{\operatorname{dec}}||^2_2, \\
        \mathcal{L} &= \frac{\mathcal{L}_{\operatorname{aux}}}{\sqrt{d_W}}  + \mathcal{L}_{\operatorname{MSE}},
    \end{aligned}
\end{equation}where $\mathbf{W}_{y} = \langle \mathbf{Y}, \mathbf{Y}^T \rangle = \mathbf{Y}\mathbf{Y}^T \in \mathbb{R}^{N \times N}$ represents the actual correlation matrix of the target future series. This auxiliary loss function compels SimBlock to align $\mathbf{W}_{\operatorname{dec}}$ with the true inter-variable relationships, ensuring that the extracted correlations are both unique and reflective of the underlying data dynamics in the future. Then this auxiliary loss will be scale with the cardinality of $d_W = \operatorname{dim}(\mathbf{W}_{y})$ and added to the common MSE loss on prediction for training. With this explicitly modeling style, SimBlock provides a prior exploration on the future dynamics as a robust foundation that constrains and guides the predictive processes of the decoder, thereby enhancing the model’s ability to forecast future sequences with greater precision and reliability.

\section{Experiments}
\label{experiments}

We conduct comprehensive empirical evaluations of the proposed Adapformer across a variety of benchmark datasets and forecasting applications to rigorously validate the performance of our framework. Furthermore, we investigate the effectiveness and generality of each proposed module when integrated into other state-of-the-art models.

\paragraph{\textbf{Datasets}} To ensure a comprehensive evaluation across various real-world domains, we include seven benchmark datasets commonly used in MTSF for our experiments. Specifically, these datasets are ETT (Electricity Transformer Temperature) with two subsets, ECL (Electricity Consuming Load), and the Weather dataset—all utilized in Informer \cite{zhou2021informer}; PeMS (Caltrans Performance Measurement System) with two subsets from SCINet \cite{liu2022scinet}; and the Solar Energy dataset from LSTNet \cite{lai2018modeling}. All datasets used in this study are publicly available. Each dataset is split into training, validation, and test sets with ratios of 70\%, 15\%, and 15\%, respectively, and the splits are made strictly in chronological order to prevent data leakage. The validation set is employed for hyperparameter tuning, while the test set is used to quantitatively evaluate the model's final performance. Lastly, all datasets are standardized based on the statistics computed from the training set. Detailed descriptions of each dataset are provided in \ref{appendix_datasets}.

\paragraph{\textbf{Baselines}} We carefully select 9 well-acknowledged time series forecasting models as our benchmark methods for performance comparison, where most of them are the current state-of-the-art from top peer-reviewed conferences. These methods are representative methods come from a diverse range of model types, including: 1) seven Transformer-based methods: iTransformer \cite{liu2023itransformer}, PatchTST \cite{nie2022time}, Crossformer \cite{zhang2023crossformer}, CARD \cite{wang2024card}, FEDformer \cite{zhou2022fedformer}, Autoformer \cite{wu2021autoformer} and vanilla Transformer \cite{vaswani2017attention}, 2) one CNN-based method: TimesNet \cite{wu2022timesnet}, and 3) one MLP-based method: DLinear \cite{zeng2023transformers}. Our model will compare against these baselines, with all models being evaluated under the same framework and environment.

\paragraph{\textbf{Implementations}} All the models and experimental frameworks are implemented entirely in Python 3.12.0~\cite{van1995python} and built upon PyTorch 2.4.0~\cite{paszke2019pyTorch}. All the experiments reported in this paper are conducted on a 16-core AMD EPYC 9654 CPU and a single NVIDIA RTX 4090 GPU. We select Adam \cite{kingma2017adam} as the optimizer with an initial learning rate in $\{5\times10^{-3},10^{-3},5\times10^{-4}\}$ and L2 loss combined with the auxiliary loss from Eq~\ref{aux_loss} to learn the model parameters. Further implementation details are presented in \ref{appendix_implementation} and the detailed hyperparameters are listed in \ref{appendix_hyperparameter}. In line with previous studies, we will assess the forecasting performance of the proposed model and all baseline methods using the widely adopted metric: the Mean Squared Error (MSE) and the Mean Absolute Error (MAE).

\subsection{Forecasting Results}

\begin{table}[H]
    \caption{Long-term Multivariate Time Series Forecasting Results. For all datasets, the lookback window length is set to $T = 96$, and we evaluate the model performance across four prediction horizons: $\{12, 24, 48, 96\}$ for the PEMS dataset and $\{96, 192, 336, 720\}$ for all other datasets. Additionally, we compare the average performance across these prediction lengths.}
    \label{table_results}
    \centering
    \resizebox{\textwidth}{!}
    {
    \begin{tabular}{cc|cc|cc|cc|cc|cc|cc|cc|cc|cc|cc}
        \toprule
        \multicolumn{2}{c|}{\multirow{2}{*}{Models}} & \multicolumn{2}{c}{\textbf{Adapformer}} & \multicolumn{2}{c}{iTransformer} & \multicolumn{2}{c}{PatchTST}  & \multicolumn{2}{c}{Crossformer}  & \multicolumn{2}{c}{CARD}  & \multicolumn{2}{c}{TimesNet}  & \multicolumn{2}{c}{DLinear}  & \multicolumn{2}{c}{FEDformer}  & \multicolumn{2}{c}{Autoformer}  & \multicolumn{2}{c}{Transformer} \\
        & & \multicolumn{2}{c}{\textbf{(Ours)}} & \multicolumn{2}{c}{(2024)} & \multicolumn{2}{c}{(2023)} & \multicolumn{2}{c}{(2023)} & \multicolumn{2}{c}{(2023)} & \multicolumn{2}{c}{(2023)} & \multicolumn{2}{c}{(2023)} & \multicolumn{2}{c}{(2023)} & \multicolumn{2}{c}{(2022)} & \multicolumn{2}{c}{(2021)} \\
        \cmidrule(lr){3-4} \cmidrule(lr){5-6} \cmidrule(lr){7-8} \cmidrule(lr){9-10} \cmidrule(lr){11-12} \cmidrule(lr){13-14} \cmidrule(lr){15-16} \cmidrule(lr){17-18} \cmidrule(lr){19-20} \cmidrule(lr){21-22}
        \multicolumn{2}{c|}{Metric} & MSE & MAE & MSE & MAE & MSE & MAE & MSE & MAE & MSE & MAE & MSE & MAE & MSE & MAE & MSE & MAE & MSE & MAE & MSE & MAE \\
        \toprule
        
        \multirow{5}{*}{\rotatebox{90}{ETTh1}} & \multicolumn{1}{|c|}{96} & \textcolor{red}{\textbf{0.423}} & \textcolor{red}{\textbf{0.442}} & 0.447 & 0.468 & \underline{\textcolor{blue}{0.426}} & \underline{\textcolor{blue}{0.457}} & \underline{\textcolor{blue}{0.426}} & 0.460 & 0.480 & 0.482 & 0.485 & 0.501 & 0.457 & 0.470 & 0.474 & 0.503 & 0.799 & 0.729 & 0.668 & 0.593 \\

        & \multicolumn{1}{|c|}{192} & \underline{\textcolor{blue}{0.469}} & \textcolor{red}{\textbf{0.485}} & 0.495 & 0.503 & \textcolor{red}{\textbf{0.466}} & \underline{\textcolor{blue}{0.489}} & 0.540 & 0.539 & 0.533 & 0.523 & 0.546 & 0.537 & 0.502 & 0.503 & 0.566 & 0.565 & 0.643 & 0.593 & 0.788 & 0.646 \\

        & \multicolumn{1}{|c|}{336} & \textcolor{red}{\textbf{0.506}} & \underline{\textcolor{blue}{0.527}} & \underline{\textcolor{blue}{0.536}} & \underline{\textcolor{blue}{0.527}} & \underline{\textcolor{blue}{0.536}} & \textcolor{red}{\textbf{0.514}} & 0.599 & 0.595 & 0.655 & 0.594 & 0.603 & 0.574 & \underline{\textcolor{blue}{0.536}} & 0.530 & 0.611 & 0.584 & 0.638 & 0.557 & 0.856 & 0.678 \\

        & \multicolumn{1}{|c|}{720} & \underline{\textcolor{blue}{0.636}} & 0.611 & 0.665 & \underline{\textcolor{blue}{0.603}} & 0.659 & 0.612 & 0.804 & 0.698 & 0.775 & 0.667 & 0.770 & 0.665 & \textcolor{red}{\textbf{0.622}} & \textcolor{red}{\textbf{0.591}} & 0.751 & 0.664 & 0.866 & 0.684 & 0.946 & 0.738 \\
        \cmidrule(lr){2-22}
        & \multicolumn{1}{|c|}{Avg} & \textcolor{red}{\textbf{0.508}} & \textcolor{red}{\textbf{0.516}} & 0.536 & 0.525 & \underline{\textcolor{blue}{0.522}} & \underline{\textcolor{blue}{0.518}} & 0.592 & 0.573 & 0.611 & 0.567 & 0.601 & 0.569 & 0.529 & 0.524 & 0.601 & 0.579 & 0.737 & 0.641 & 0.815 & 0.664 \\
        \midrule

        \multirow{5}{*}{\rotatebox{90}{ETTh2}} & \multicolumn{1}{|c|}{96} & \underline{\textcolor{blue}{0.150}} & \underline{\textcolor{blue}{0.273}} & 0.162 & 0.283 & \textcolor{red}{\textbf{0.149}} & \textcolor{red}{\textbf{0.269}} & 0.170 & 0.311 & 0.166 & 0.285 & 0.167 & 0.289 & 0.151 & 0.274 & 0.180 & 0.303 & 0.195 & 0.326 & 0.198 & 0.336 \\

        & \multicolumn{1}{|c|}{192} & \underline{\textcolor{blue}{0.187}} & 0.308 & \underline{\textcolor{blue}{0.187}} & \underline{\textcolor{blue}{0.303}} & \textcolor{red}{\textbf{0.182}} & \textcolor{red}{\textbf{0.298}} & 0.191 & 0.334 & 0.188 & 0.305 & 0.210 & 0.323 & \textcolor{red}{\textbf{0.182}} & 0.304 & 0.199 & 0.318 & 0.212 & 0.333 & 0.268 & 0.386 \\

        & \multicolumn{1}{|c|}{336} & 0.209 & 0.323 & 0.205 & \underline{\textcolor{blue}{0.318}} & \underline{\textcolor{blue}{0.204}} & \textcolor{red}{\textbf{0.315}} & 0.263 & 0.385 & 0.219 & 0.328 & 0.216 & 0.330 & \textcolor{red}{\textbf{0.203}} & 0.327 & 0.212 & 0.332 & 0.240 & 0.363 & 0.259 & 0.406 \\

        & \multicolumn{1}{|c|}{720} & \textcolor{red}{\textbf{0.244}} & \underline{\textcolor{blue}{0.356}} & 0.253 & \textcolor{red}{\textbf{0.355}} & 0.253 & \textcolor{red}{\textbf{0.355}} & 0.519 & 0.491 & 0.272 & 0.370 & \underline{\textcolor{blue}{0.248}} & \textcolor{red}{\textbf{0.355}} & 0.266 & 0.378 & 0.259 & 0.367 & 0.505 & 0.495 & 0.404 & 0.472 \\
        \cmidrule(lr){2-22}
        & \multicolumn{1}{|c|}{Avg} & \textcolor{red}{\textbf{0.197}} & \underline{\textcolor{blue}{0.315}} & 0.202 & \underline{\textcolor{blue}{0.315}} & \textcolor{red}{\textbf{0.197}} & \textcolor{red}{\textbf{0.309}} & 0.286 & 0.380 & 0.211 & 0.322 & 0.210 & 0.324 & \underline{\textcolor{blue}{0.201}} & 0.321 & 0.212 & 0.330 & 0.288 & 0.379 & 0.282 & 0.400 \\
        \midrule

        \multirow{5}{*}{\rotatebox{90}{ECL}} & \multicolumn{1}{|c|}{96} & \textcolor{red}{\textbf{0.143}} & \underline{\textcolor{blue}{0.242}} & \underline{\textcolor{blue}{0.149}} & \textcolor{red}{\textbf{0.239}} & 0.181 & 0.266 & 0.165 & 0.266 & 0.199 & 0.277 & 0.168 & 0.270 & 0.211 & 0.300 & 0.209 & 0.317 & 0.200 & 0.312 & 0.453 & 0.484 \\

        & \multicolumn{1}{|c|}{192} & \textcolor{red}{\textbf{0.159}} & \underline{\textcolor{blue}{0.251}} & \underline{\textcolor{blue}{0.160}} & \textcolor{red}{\textbf{0.248}} & 0.183 & 0.272 & 0.183 & 0.283 & 0.197 & 0.280 & 0.182 & 0.283 & 0.207 & 0.301 & 0.219 & 0.328 & 0.222 & 0.330 & 0.382 & 0.430 \\

        & \multicolumn{1}{|c|}{336} & \textcolor{red}{\textbf{0.174}} & \textcolor{red}{\textbf{0.262}} & \underline{\textcolor{blue}{0.177}} & \underline{\textcolor{blue}{0.263}} & 0.196 & 0.285 & 0.214 & 0.308 & 0.205 & 0.288 & 0.199 & 0.298 & 0.217 & 0.314 & 0.221 & 0.332 & 0.234 & 0.339 & 0.518 & 0.534 \\

        & \multicolumn{1}{|c|}{720} & \underline{\textcolor{blue}{0.211}} & \textcolor{red}{\textbf{0.281}} & \textcolor{red}{\textbf{0.208}} & \underline{\textcolor{blue}{0.290}} & 0.234 & 0.314 & 0.285 & 0.376 & 0.242 & 0.319 & 0.218 & 0.315 & 0.248 & 0.344 & 0.273 & 0.374 & 0.263 & 0.365 & 0.366 & 0.425 \\
        \cmidrule(lr){2-22}
        & \multicolumn{1}{|c|}{Avg} & \textcolor{red}{\textbf{0.172}} & \textcolor{red}{\textbf{0.259}} & \underline{\textcolor{blue}{0.173}} & \underline{\textcolor{blue}{0.260}} & 0.199 & 0.284 & 0.211 & 0.308 & 0.211 & 0.291 & 0.192 & 0.291 & 0.221 & 0.315 & 0.231 & 0.338 & 0.230 & 0.337 & 0.430 & 0.468 \\
        \midrule

        \multirow{5}{*}{\rotatebox{90}{Weather}} & \multicolumn{1}{|c|}{96} & \textcolor{red}{\textbf{0.144}} & \textcolor{red}{\textbf{0.195}} & 0.166 & \underline{\textcolor{blue}{0.204}} & 0.164 & \underline{\textcolor{blue}{0.204}} & \underline{\textcolor{blue}{0.163}} & 0.230 & 0.174 & 0.218 & 0.166 & 0.214 & 0.185 & 0.244 & 0.198 & 0.283 & 0.240 & 0.349 & 0.706 & 0.606 \\

        & \multicolumn{1}{|c|}{192} & \underline{\textcolor{blue}{0.208}} & 0.260 & \underline{\textcolor{blue}{0.208}} & \underline{\textcolor{blue}{0.242}} & \textcolor{red}{\textbf{0.205}} & \textcolor{red}{\textbf{0.239}} & 0.223 & 0.294 & 0.217 & 0.255 & \underline{\textcolor{blue}{0.208}} & 0.249 & 0.225 & 0.284 & 0.281 & 0.353 & 0.303 & 0.352 & 1.221 & 0.893 \\

        & \multicolumn{1}{|c|}{336} & \textcolor{red}{\textbf{0.256}} & \textcolor{red}{\textbf{0.273}} & \underline{\textcolor{blue}{0.257}} & 0.280 & \textcolor{red}{\textbf{0.256}} & \underline{\textcolor{blue}{0.279}} & 0.258 & 0.320 & 0.271 & 0.296 & 0.271 & 0.292 & 0.268 & 0.320 & 0.358 & 0.394 & 0.315  & 0.373 & 0.884 & 0.732 \\

        & \multicolumn{1}{|c|}{720} & \textcolor{red}{\textbf{0.315}} & \textcolor{red}{\textbf{0.320}} & \underline{\textcolor{blue}{0.316}} & \underline{\textcolor{blue}{0.322}} & \underline{\textcolor{blue}{0.316}} & \textcolor{red}{\textbf{0.320}} & 0.325 & 0.376 & 0.323 & 0.329 & 0.322 & 0.327 & 0.328 & 0.374 & 0.373 & 0.398 & 0.399 & 0.427 & 0.892 & 0.748 \\
        \cmidrule(lr){2-22}
        & \multicolumn{1}{|c|}{Avg} & \textcolor{red}{\textbf{0.230}} & \underline{\textcolor{blue}{0.262}} & 0.237 & \underline{\textcolor{blue}{0.262}} & \underline{\textcolor{blue}{0.235}} & \textcolor{red}{\textbf{0.261}} & 0.242 & 0.305 & 0.246 & 0.274 & 0.242 & 0.270 & 0.252 & 0.305 & 0.302 & 0.357 & 0.314 & 0.375 & 0.926 & 0.745 \\
        \midrule

         \multirow{5}{*}{\rotatebox{90}{PEMS03}} & \multicolumn{1}{|c|}{12} & \underline{\textcolor{blue}{0.068}} & \textcolor{red}{\textbf{0.171}} & \textcolor{red}{\textbf{0.065}} & \textcolor{red}{\textbf{0.171}} & 0.088 & 0.206 & 0.083 & 0.194 & -  & - & 0.084 & \underline{\textcolor{blue}{0.191}} & 0.105 & 0.229 & 0.119 & 0.246 & 0.295 & 0.406 & 0.108 & 0.207 \\

        & \multicolumn{1}{|c|}{24} & \underline{\textcolor{blue}{0.094}} & \textcolor{red}{\textbf{0.204}} & \textcolor{red}{\textbf{0.093}} & \underline{\textcolor{blue}{0.206}} & 0.144 & 0.263 & 0.117 & 0.238 & - & - & 0.114 & 0.221 & 0.177 & 0.302 & 0.144 & 0.274 & 0.382 & 0.467 & 0.126 & 0.230 \\

        & \multicolumn{1}{|c|}{48} & \textcolor{red}{\textbf{0.149}} & \textcolor{red}{\textbf{0.259}} & \underline{\textcolor{blue}{0.161}} & 0.274 & 0.275 & 0.367 & 0.205 & 0.319 & - & - & 0.176 & 0.273 & 0.308 & 0.413 & 0.229 & 0.355 & 0.685 & 0.628 & 0.164 & \underline{\textcolor{blue}{0.265}} \\

        & \multicolumn{1}{|c|}{96} & \textcolor{red}{\textbf{0.186}} & \underline{\textcolor{blue}{0.301}} & 0.348 & 0.456 & 0.508 & 0.528 & 0.259 & 0.363 & - & - & 0.290 & 0.342 & 0.448 & 0.513 & 0.364 & 0.456 & 0.908 & 0.747 & \underline{\textcolor{blue}{0.189}} & \textcolor{red}{\textbf{0.283}} \\
        \cmidrule(lr){2-22}
        & \multicolumn{1}{|c|}{Avg} & \textcolor{red}{\textbf{0.124}} & \textcolor{red}{\textbf{0.234}} & 0.167 & 0.277 & 0.254 & 0.341 & 0.166 & 0.278 & - & - & 0.166 & 0.257 & 0.260 & 0.364 & 0.214 & 0.333 & 0.568 & 0.562 & \underline{\textcolor{blue}{0.147}} & \underline{\textcolor{blue}{0.246}} \\
        \midrule

        \multirow{5}{*}{\rotatebox{90}{PEMS07}} & \multicolumn{1}{|c|}{12} & \textcolor{red}{\textbf{0.065}} & \textcolor{red}{\textbf{0.164}} & \underline{\textcolor{blue}{0.067}} & \underline{\textcolor{blue}{0.166}} & 0.091 & 0.209 & 0.095 & 0.199 & -  & - & 0.105 & 0.222 & 0.110 & 0.231 & 0.109 & 0.224 & 0.245 & 0.360 & 0.166 & 0.229 \\

        & \multicolumn{1}{|c|}{24} & \textcolor{red}{\textbf{0.089}} & \textcolor{red}{\textbf{0.193}} & \underline{\textcolor{blue}{0.098}} & \underline{\textcolor{blue}{0.201}} & 0.153 & 0.270 & 0.143 & 0.254 & - & - & 0.130 & 0.248 & 0.204 & 0.319 & 0.125 & 0.242 & 0.280 & 0.388 & 0.175 & 0.240 \\

        & \multicolumn{1}{|c|}{48} & \textcolor{red}{\textbf{0.127}} & \textcolor{red}{\textbf{0.234}} & 0.179 & 0.292 & 0.303 & 0.382 & 0.256 & 0.331 & - & - & 0.178 & 0.294 & 0.394 & 0.450 & \underline{\textcolor{blue}{0.173}} & 0.295 & 0.265 & 0.380 & 0.181 & \underline{\textcolor{blue}{0.248}} \\

        & \multicolumn{1}{|c|}{96} & \textcolor{red}{\textbf{0.167}} & \underline{\textcolor{blue}{0.270}} & 0.392 & 0.554 & 0.561 & 0.541 & 0.290 & 0.372 & - & - & 0.255 & 0.356 & 0.598 & 0.549 & 0.275 & 0.387 & 0.464 & 0.510 & \underline{\textcolor{blue}{0.186}} & \textcolor{red}{\textbf{0.253}} \\
        \cmidrule(lr){2-22}
        & \multicolumn{1}{|c|}{Avg} & \textcolor{red}{\textbf{0.112}} & \textcolor{red}{\textbf{0.215}} & 0.184 & 0.303 & 0.277 & 0.351 & 0.196 & 0.289 & - & - & \underline{\textcolor{blue}{0.167}} & 0.280 & 0.327 & 0.387 & 0.170 & 0.287 & 0.314 & 0.410 & 0.177 & \underline{\textcolor{blue}{0.242}} \\
        \midrule

        \multirow{5}{*}{\rotatebox{90}{Solar}} & \multicolumn{1}{|c|}{96} & \textcolor{red}{\textbf{0.170}} & \textcolor{red}{\textbf{0.235}} & 0.211 & \underline{\textcolor{blue}{0.247}} & 0.215 & 0.276 & \underline{\textcolor{blue}{0.193}} & 0.257 & 0.266  & 0.295 & 0.220 & 0.237 & 0.280 & 0.374 & 0.268 & 0.378 & 0.507 & 0.512 & 0.735 & 0.769 \\

        & \multicolumn{1}{|c|}{192} & \textcolor{red}{\textbf{0.225}} & \underline{\textcolor{blue}{0.272}} & 0.239 & \textcolor{red}{\textbf{0.271}} & 0.240 & 0.283 & \underline{\textcolor{blue}{0.230}} & 0.281 & 0.280 & 0.300 & 0.248 & 0.282 & 0.310 & 0.394 & 0.335 & 0.418 & 0.688 & 0.611 & 0.733 & 0.751 \\

        & \multicolumn{1}{|c|}{336} & \textcolor{red}{\textbf{0.237}} & \textcolor{red}{\textbf{0.276}} & 0.258 & 0.285 & 0.263 & 0.300 & \underline{\textcolor{blue}{0.251}} & 0.300 & 0.311 & 0.321 & \underline{\textcolor{blue}{0.282}} & 0.282 & 0.338 & 0.409 & 0.283 & 0.373 & 0.752 & 0.647 & 0.757 & 0.778 \\

        & \multicolumn{1}{|c|}{720} & \textcolor{red}{\textbf{0.191}} & \textcolor{red}{\textbf{0.252}} & 0.255 & \underline{\textcolor{blue}{0.285}} & \underline{\textcolor{blue}{0.249}} & 0.292 & 0.514 & 0.533 & 0.316 & 0.318 & 0.274 & 0.295 & 0.335 & 0.403 & 0.367 & 0.440 & 0.793 & 0.698 & 0.760 & 0.781 \\
        \cmidrule(lr){2-22}
        & \multicolumn{1}{|c|}{Avg} & \textcolor{red}{\textbf{0.206}} & \textcolor{red}{\textbf{0.259}} & \underline{\textcolor{blue}{0.241}} & \underline{\textcolor{blue}{0.272}} & 0.242 & 0.288 & 0.297 & 0.343 & 0.293 & 0.308 & 0.256 & 0.274 & 0.316 & 0.395 & 0.313 & 0.402 & 0.685 & 0.617 & 0.746 & 0.770 \\
        \midrule

        \multicolumn{2}{c|}{1\textsuperscript{st} Count} & \textcolor{red}{\textbf{26}} & \textcolor{red}{\textbf{21}} & 3 & 5 & \underline{\textcolor{blue}{6}} & \underline{\textcolor{blue}{9}} & 0 & 0 & 0 & 0 & 0 & 1 & 3 & 1 & 0 & 0 & 0 & 0 & 0 & 2 \\

        \cmidrule{3-22}
        \multicolumn{2}{c|}{Top 2 Count} & \textcolor{red}{\textbf{34}} & \textcolor{red}{\textbf{29}} & \underline{\textcolor{blue}{16}} & \underline{\textcolor{blue}{23}} & 13 & 14 & 5 & 0 & 0 & 0 & 4 & 2 & 5 & 1 & 1 & 0 & 0 & 0 & 3 & 6\\

        \bottomrule
    \end{tabular}
    }
\end{table}

The complete results of multivariate time series forecasting for our proposed Adapformer model and the baseline models across seven datasets are presented in Table \ref{table_results}. In the table, the best results are highlighted in \textbf{\textcolor{red}{bold}}, and the second-best are \underline{\textcolor{blue}{underlined}}, while "-" indicates a model ran out of memory. The results clearly demonstrate that Adapformer consistently achieves lower MSE and MAE values across the majority of datasets. Noticeably, out of 35 reported MSE results, Adapformer ranks within the top two in 34 instances and secures the top position in 26, outperforming the prior state-of-the-art models and establishing a new benchmark for forecasting accuracy. This strong performance underscores the efficacy of our two-stage design.

It is notable that our approach preforms outstanding on both PEMS dataset, which are characterized by large amount of variables and hence possess complicated and essential cross-variate relationships as well as interferences. Our average lead in MSE over the second place on both datasets, PEMS03 and PEMS07, is 15.6\% and 32.9\% respectively. Such performance demonstrates the superiority of Adapformer when confronted with tremendous exogenous noise brought by increasing amount of interfering variables. 

Moreover, existing state-of-the-art models can be systematically categorized into two distinct groups: channel-independent (CI) approaches, represented by PatchTST and DLinear, and channel-dependent (CD) approaches, exemplified by iTransformer, Crossformer, and TimesNet. This classification allows for a comprehensive comparative analysis of their respective performances. Our empirical results demonstrate that CD approaches significantly outperform CI methods on both the PEMS03 and PEMS07 datasets. This pronounced superiority underscores the critical importance of capturing and leveraging inter-variable dependencies, which are essential for comprehensively understanding the latent statistical properties. Such governs the intricate interactions under large dimensionality scale. In contrast, CI approaches, which treat each variate as separate individuals, inherently lack the ability to access any of these inter-dependencies when facing large amount of variables. Conversely, channel-independent (CI) approaches have demonstrated superior performance on ETT datasets, which are characterized by a relatively limited number of variables. In such contexts, each variate tends to encapsulate distinct spatio-temporal properties independently, thereby aligning well with the inherent strengths of CI methodologies. This independence allows CI models to effectively capture and model the unique dependencies within each variate, making them particularly well-suited for datasets with fewer channels. The strengths and weaknesses of both types of approaches, as discussed, are empirically validated. Building on these insights, our Adapformer outperforms all baseline methods, indicating the success of our adopted channel management strategy.  To enhance the model's capacity, our Adapformer embodies the integrative strategy by emphasizing the learning of CI information within the CD framework. This represents a breakthrough beyond the two basic frameworks. By incorporating CI learning mechanisms, just like our Adaptive Channel Enhancer in the Adapformer framework, the model avoids the pitfalls of focusing solely on inter-variable interactions, thereby maintaining the integrity of each channel's intrinsic properties. Empirical results further support the efficacy of Adapformer, demonstrating its ability to achieve superior forecasting performance by effectively mitigating noise while harnessing both CI and CD information. In addition to the quantitative results, qualitative comparisons can be found in \ref{appendix_showcase} to further demonstrate the ability of Adapformer to closely track temporal patterns and align with ground truth.

In addition, as time-series forecasting typically prioritizes minimizing absolute errors so that predictions remain accurate in their original units, we have so far focused on MSE and MAE. However, to provide a complementary perspective on each model’s ability to capture the underlying variance in the data, we also report the coefficient of determination ($R^2$) in Table \ref{table_r}. Unlike MSE and MAE, which measure the average magnitude of residuals, $R^2$ quantifies the proportion of target variance that the model explains, offering a relative “goodness-of-fit” view. That said, $R^2$ has its own limitations: it can be sensitive to non-stationarity or heteroscedasticity in the series, may become negative when models perform worse than a constant baseline, and does not reflect the scale of forecasting errors. For these reasons, we present $R^2$ as a supplementary metric, using it alongside MSE and MAE to give a more rounded evaluation of forecasting performance. Here, Adapformer’s unusually flat $R^2$ decline indicates it is not merely minimizing residuals, but actually learning and preserving the dominant temporal patterns even as noise builds up over longer horizons. In contrast, channel‐dependent methods such as Crossformer often ``overfit” on short‐term cross‐channel signals, achieving high $R^2$ initially but only to see their explanatory power collapse when correlations shift or weaken. Such scenario is common to observe especially when noise and non-stationarity accumulate over larger prediction length. Purely channel‐independent approaches such as PatchTST and DLinear suffer less from this volatility but never reach the same peak $R^2$, underscoring their difficulty in leveraging inter‐series information. These dynamics suggest that the most robust forecasters should combine both strengths: balancing absolute accuracy with genuine variance capture.

\begin{table}
    
    \centering
    \caption{$R^2$ comparison of Adapformer against nine baseline models on four datasets (ETTh1, ETTh2, Weather, PEMS07), with average R² reported for each dataset.}
    \label{table_r}
    \resizebox{\textwidth}{!}
    {
    \small
    \begin{tabular}{l|c|c|c|c|c|c|c|c|c|c|c}
        \toprule
        \multicolumn{2}{c|}{Metric} & \multicolumn{10}{c}{R\textsuperscript{2}} \\
        \cmidrule(lr){3-12}
        \multicolumn{2}{c|}{Models} & \makebox[6em][c]{Adapformer} & \makebox[6em][c]{iTransformer} & \makebox[6em][c]{PatchTST} & \makebox[6em][c]{Crossformer} & \makebox[6em][c]{CARD} & \makebox[6em][c]{TimesNet} & \makebox[6em][c]{DLinear} & \makebox[6em][c]{FEDformer} & \makebox[6em][c]{Autoformer} & \makebox[6em][c]{Transformer} \\
        \midrule
        
        \multirow{5}{*}{\rotatebox{90}{ETTh1}}
        & 96  & \underline{\textcolor{blue}{0.614}} & 0.603 & 0.547 & \textcolor{red}{\textbf{0.628}} & 0.567 & 0.572 & 0.601 & 0.590 & 0.312 & 0.416 \\
        & 192 & \textcolor{red}{\textbf{0.588}} & \underline{\textcolor{blue}{0.582}} & 0.531 & 0.568 & 0.546 & 0.542 & 0.579 & 0.521 & 0.461 & 0.315 \\
        & 336 & \underline{\textcolor{blue}{0.554}} & \underline{\textcolor{blue}{0.554}} & 0.516 & 0.500 & 0.454 & 0.495 & \textcolor{red}{\textbf{0.557}} & 0.526 & 0.472 & 0.270 \\
        & 720 & \underline{\textcolor{blue}{0.464}} & 0.437 & 0.435 & 0.273 & 0.359 & 0.373 & \textcolor{red}{\textbf{0.483}} & 0.377 & 0.280 & 0.208 \\
        \cmidrule{2-12}
        & Avg & \textcolor{red}{\textbf{0.555}} & \underline{\textcolor{blue}{0.544}} & 0.507 & 0.492 & 0.481 & 0.495 & \textcolor{red}{\textbf{0.555}} & 0.504 & 0.381 & 0.302 \\
        \midrule

        \multirow{5}{*}{\rotatebox{90}{ETTh2}}
        & 96  & \underline{\textcolor{blue}{0.717}} & 0.696 & 0.682 & 0.667 & 0.692 & 0.691 & \textcolor{red}{\textbf{0.720}} & 0.667 & 0.638 & 0.600 \\
        & 192 & 0.646 & 0.635 & 0.649 & 0.646 & \underline{\textcolor{blue}{0.661}} & 0.625 & \textcolor{red}{\textbf{0.672}} & 0.638 & 0.619 & 0.502 \\
        & 336 & \underline{\textcolor{blue}{0.634}} & 0.624 & 0.630 & 0.534 & 0.613 & 0.616 & \textcolor{red}{\textbf{0.639}} & 0.620 & 0.576 & 0.478 \\
        & 720 & \underline{\textcolor{blue}{0.555}} & \underline{\textcolor{blue}{0.555}} & \textcolor{red}{\textbf{0.583}} & 0.286 & 0.525 & 0.544 & 0.536 & 0.548 & 0.370 & 0.332 \\
        \cmidrule{2-12}
        & Avg & \underline{\textcolor{blue}{0.638}} & 0.627 & 0.636 & 0.533 & 0.623 & 0.619 & \textcolor{red}{\textbf{0.642}} & 0.618 & 0.551 & 0.478 \\
        \midrule

        \multirow{5}{*}{\rotatebox{90}{Weather}}
        & 96  & \textcolor{red}{\textbf{0.718}} & 0.681 & 0.671 & \underline{\textcolor{blue}{0.685}} & 0.657 & 0.678 & 0.650 & 0.612 & 0.475 & 0.378 \\
        & 192 & \textcolor{red}{\textbf{0.621}} & 0.601 & 0.594 & 0.576 & 0.584 & \underline{\textcolor{blue}{0.603}} & 0.587 & 0.450 & 0.286 & 0.113 \\
        & 336 & \textcolor{red}{\textbf{0.548}} & 0.522 & 0.522 & 0.521 & 0.495 & 0.505 & \underline{\textcolor{blue}{0.525}} & 0.363 & 0.379 & 0.280 \\
        & 720 & 0.420 & 0.413 & \textcolor{red}{\textbf{0.427}} & 0.406 & 0.399 & 0.404 & \underline{\textcolor{blue}{0.425}} & 0.317 & 0.239 & 0.209 \\
        \cmidrule{2-12}
        & Avg & \textcolor{red}{\textbf{0.577}} & 0.544 & \underline{\textcolor{blue}{0.553}} & 0.547 & 0.534 & 0.547 & 0.547 & 0.435 & 0.345 & 0.245 \\
        \midrule

        \multirow{5}{*}{\rotatebox{90}{PEMS07}}
        & 12 & \textcolor{red}{\textbf{0.764}} & \underline{\textcolor{blue}{0.752}} & 0.242 & 0.643 & - & 0.589 & 0.592 & 0.655 & 0.630 & 0.394 \\
        & 24 & \underline{\textcolor{blue}{0.702}} & \textcolor{red}{\textbf{0.730}} & 0.233 & 0.509 & - & 0.528 & 0.573 & 0.507 & 0.634 & 0.408 \\
        & 48 & \textcolor{red}{\textbf{0.649}} & \underline{\textcolor{blue}{0.610}} & 0.273 & 0.348 & - & 0.463 & 0.509 & 0.423 & 0.444 & 0.460 \\
        & 96 & \textcolor{red}{\textbf{0.664}} & 0.441 & 0.360 & 0.389 & - & 0.501 & 0.477 & 0.379 & 0.167 & \underline{\textcolor{blue}{0.600}} \\
        \cmidrule{2-12}
        & Avg & \textcolor{red}{\textbf{0.695}} & \underline{\textcolor{blue}{0.633}} & 0.277 & 0.472 & - & 0.520 & 0.538 & 0.491 & 0.469 & 0.466 \\
        \bottomrule
    \end{tabular}
    }
\end{table}

\subsection{Ablation Study}
\label{subsection_ablation_study}

To rigorously evaluate the individual contributions of the core components within our model design, we conduct an ablation study focusing on three pivotal modules: the Adaptive Channel Enhancer (ACE), the Adaptive Channel Forecaster (ACF), and the auxiliary loss function for the SimBlock. This study involves systematically removing each of these modules in isolation to assess their impact on the model's overall forecasting performance. Specifically, we analyze the performance degradation when \textbf{1.} \textbf{ACE} is omitted, thereby examining the role of adaptive channel management in integrating inter-variable dependencies; \textbf{2.} \textbf{ACF} is replaced by simple MLP, to evaluate the significance of adaptive feature extraction in interpreting channel-specific information and mitigating noise during predicting stage; \textbf{3.} \textbf{Auxiliary loss} is removed, to understand its influence on the model’s optimization and generalization capabilities.

\begin{table}[H]
    \caption{Ablation Study of Adapformer. This table displays the detailed performance of our model on three datasets (ETTh1, Weather and Solar) with either ACE, ACF or the auxiliary loss being removed. In the table, a $\downarrow$ means a degradation in model performance, a $\uparrow$ means a promotion in model performance and a $\mbox{--}$ means an even in model performance.}
    \label{table_ablation}
    \centering
    \resizebox{\textwidth}{!}
    {
        \begin{tabular}{|c|c|cccc|c|cccc|c|cccc|c|}
            \toprule
            \multicolumn{2}{|c|}{Dataset} & \multicolumn{5}{c|}{ETTh1} & \multicolumn{5}{c|}{Weather} & \multicolumn{5}{c|}{Solar} \\
    
            \cmidrule{3-17}
    
            \multicolumn{2}{|c|}{Models} & 96 & 192 & 336 & 720 & Avg & 96 & 192 & 336 & 720 & Avg & 96 & 192 & 336 & 720 & Avg \\
    
            \midrule
    
            \multirow{2}{*}{Original} & MSE & \textbf{0.423} & \textbf{0.469} & \textbf{0.506} & \textbf{0.636} & \textbf{0.508} & \textbf{0.144} & \textbf{0.208} & \textbf{0.256} & \textbf{0.315} & \textbf{0.230} & \textbf{0.170} & \textbf{0.225} & \textbf{0.237} & \textbf{0.228} & \textbf{0.215} \\
    
            \cmidrule{2-17}
    
             \rowcolor{black!10} \cellcolor{white} & \cellcolor{white} Performance & \textbf{$\mbox{--}$} & \textbf{$\mbox{--}$} & \textbf{$\mbox{--}$} & \textbf{$\mbox{--}$} & \textbf{100\%} & \textbf{$\mbox{--}$} & \textbf{$\mbox{--}$} & \textbf{$\mbox{--}$} & \textbf{$\mbox{--}$} & \textbf{100\%} & \textbf{$\mbox{--}$} & \textbf{$\mbox{--}$} & \textbf{$\mbox{--}$} & \textbf{$\mbox{--}$} & \textbf{100\%} \\
    
             \midrule
    
             \multirow{2}{*}{\textit{w/o} ACE} & MSE & 0.425 & 0.474 & 0.528 & 0.641 & 0.516 & 0.153 & 0.210 & 0.256 & 0.329 & 0.233 & 0.218 & 0.237 & 0.241 & 0.240 & 0.233 \\
    
            \cmidrule{2-17}
    
             \rowcolor{black!10} \cellcolor{white} & \cellcolor{white} Performance & $\downarrow$ & $\downarrow$ & $\downarrow$ & $\downarrow$ & 98.4\% & $\downarrow$ & $\downarrow$ & $\mbox{--}$ & $\downarrow$ & 99.1\% & $\downarrow$ & $\downarrow$ & $\downarrow$ & $\downarrow$ & 94.8\% \\
    
             \midrule
    
             \multirow{2}{*}{\textit{w/o} ACF} & MSE & 0.438 & 0.489 & 0.551 & 0.711 & 0.547 & 0.149 & 0.221 & 0.264 & 0.322 & 0.239 & 0.225 & 0.236 & 0.242 & 0.249 & 0.238 \\
    
            \cmidrule{2-17}
    
             \rowcolor{black!10} \cellcolor{white} & \cellcolor{white} Performance & $\downarrow$ & $\downarrow$ & $\downarrow$ & $\downarrow$ & 92.8\% & $\downarrow$ & $\downarrow$ & $\downarrow$ & $\downarrow$ & 96.6\% & $\downarrow$ & $\downarrow$ & $\downarrow$ & $\downarrow$ & 95.7\% \\

             \midrule
    
             \multirow{2}{*}{\textit{w/o} AUX} & MSE & 0.427 & 0.529 & 0.506 & 0.636 & 0.524 & 0.153 & 0.208 & 0.256 & 0.322 & 0.235 & 0.232 & 0.235 & 0.237 & 0.263 & 0.242 \\
    
            \cmidrule{2-17}
    
             \rowcolor{black!10} \cellcolor{white} & \cellcolor{white} Performance & $\downarrow$ & $\downarrow$ & $\mbox{--}$ & $\mbox{--}$ & 96.9\% & $\downarrow$ & $\mbox{--}$ & $\mbox{--}$ & $\downarrow$ & 98.2\% & $\downarrow$ & $\downarrow$ & $\mbox{--}$ & $\downarrow$ & 91.3\% \\
        
            \bottomrule
        \end{tabular}
    }
\end{table}

Table \ref{table_ablation} gives the full results of our ablation study. We specifically chose three datasets: ETTh1, Weather and Solar to test.  It is evident that removing each module results in a noticeable degradation in prediction performance, with the most substantial decline observed when the ACF module is excluded. This highlights the effectiveness of this novel channel management strategy when applied in the predicting stage, demonstrating Adapformer's exceptional ability to balance efficient information utilization with noise interference mitigation. Specifically, the Solar dataset contains 137 variables, but utilizing only 3-5 relevant covariates for each target can significantly enhance the prediction performance. Furthermore, the marked performance deterioration on the Solar dataset upon the removal of the ACE module underscores the superiority of our approach in selectively incorporating channel-specific temporal key points into the token representations. These findings validate the critical role of both ACE and ACF in optimizing Adapformer's performance, particularly in high-dimensional environments where selective channel management is essential for maintaining forecasting accuracy and robustness. Moreover, the slight degradation on performance when the Auxiliary loss is not employed demonstrates the necessity on constraining and supervising the learning towards the future inter-series relationships.

\subsection{Module Generalizability}

Moreover, our proposed ACE and ACF modules are intentionally designed to be lightweight and modular, allowing them to be seamlessly integrated into a wide range of Transformer-based architectures. To verify the generalizability and effectiveness of each module, we conduct two sets of experiments. Specifically, we individually incorporate the ACE and ACF modules into several strong baseline models, including iTransformer, PatchTST, and vanilla Transformer. The results, summarized in Table \ref{table_ace} and Table \ref{table_acf}, demonstrate that both modules consistently improve performance across diverse datasets and backbone architectures, validating their plug-and-play design and broad applicability.

\begin{table*}
    \caption{
    Forecasting performance of iTransformer, PatchTST, Crossformer, and Transformer before and after integrating the Adaptive Channel Enhancer (ACE) module. Results are reported in MSE and MAE on four datasets: ETTh2, ECL, Weather, and PEMS03, following the same experimental settings as our main results without additional hyperparameter tuning. The “Promotion” row shows the relative improvements (\%) brought by ACE, calculated based on the original averaged MSE and MAE.}

    \label{table_ace}
    \begin{center}\resizebox{\textwidth}{!}{
    \begin{tabular}{cc|cc>{\columncolor{shadow}}c>{\columncolor{shadow}}c|cc>{\columncolor{shadow}}c>{\columncolor{shadow}}c|cc>{\columncolor{shadow}}c>{\columncolor{shadow}}c|cc>{\columncolor{shadow}}c>{\columncolor{shadow}}c}
        \toprule
         \multicolumn{2}{c|}{\multirow{2}{*}{Models}} & \multicolumn{2}{c}{\multirow{2}{*}{iTransformer}} & \multicolumn{2}{c|}{\textbf{w/ ACE}} & \multicolumn{2}{c}{\multirow{2}{*}{PatchTST}} & \multicolumn{2}{c|}{\textbf{w/ ACE}} & \multicolumn{2}{c}{\multirow{2}{*}{Crossformer}} & \multicolumn{2}{c|}{\textbf{w/ ACE}} & \multicolumn{2}{c}{\multirow{2}{*}{Transformer}} & \multicolumn{2}{c}{\textbf{w/ ACE}}\\
        & \multicolumn{3}{c}{} & \multicolumn{2}{c|}{\textbf{(Ours)}} & \multicolumn{2}{c}{} & \multicolumn{2}{c|}{\textbf{(Ours)}} & \multicolumn{2}{c}{} & \multicolumn{2}{c|}{\textbf{(Ours)}} & \multicolumn{2}{c}{} & \multicolumn{2}{c}{\textbf{(Ours)}}\\
        \cmidrule(lr){3-6} \cmidrule(lr){7-10} \cmidrule(lr){11-14} \cmidrule(lr){15-18}
        \multicolumn{2}{c|}{Metric} & MSE & MAE & MSE & MAE & MSE & MAE & MSE & MAE & MSE & MAE & MSE & MAE & MSE & MAE & MSE & MAE  \\
        \specialrule{0.75pt}{0.0pt}{1.5pt}
        \multirow{5}{*}{\rotatebox{90}{ETTh2}} & \multicolumn{1}{|c|}{96} & 0.162 & 0.283 & \textbf{0.155} & \textbf{0.276} & 0.149 & 0.269 & \textbf{0.147} & \textbf{0.268} & 0.170 & 0.311 & \textbf{0.168} & \textbf{0.295} & 0.198 & 0.336 & \textbf{0.188} & \textbf{0.334}\\
         & \multicolumn{1}{|c|}{192}& 0.187 & 0.303 & \textbf{0.185} & \textbf{0.301} & 0.182 & 0.298 & \textbf{0.181} & \textbf{0.298} & 0.191 & 0.334 & \textbf{0.190} & \textbf{0.329} & 0.268 & 0.386 & \textbf{0.242} & \textbf{0.366} \\
         & \multicolumn{1}{|c|}{336}& 0.205 & 0.318 & \textbf{0.200} & \textbf{0.317} & 0.204 & 0.315 & \textbf{0.201} & \textbf{0.313} & 0.263 & 0.385 & \textbf{0.236} & \textbf{0.361} & 0.259 & 0.406 & \textbf{0.247} & 0.408  \\
         & \multicolumn{1}{|c|}{720}& 0.253 & 0.355 & \textbf{0.251} & 0.357 & 0.253 & 0.355 & \textbf{0.246} & \textbf{0.349} & 0.519 & 0.491 & \textbf{0.413} & 0.493 & 0.404 & 0.472 & \textbf{0.383} & \textbf{0.460} \\
         \cmidrule(lr){2-18}
         & \multicolumn{1}{|c|}{Avg} & 0.202 & 0.315 & \textbf{0.198} & \textbf{0.313} & 0.197 & 0.309 & \textbf{0.194} & \textbf{0.307} & 0.286 & 0.380 & \textbf{0.252} & \textbf{0.370} & 0.282 & 0.400 & \textbf{0.265} & \textbf{0.392} \\
        
        \specialrule{0.5pt}{1.5pt}{1.5pt}
        \multirow{5}{*}{\rotatebox{90}{ECL}} & \multicolumn{1}{|c|}{96} & 0.149 & 0.239 & \textbf{0.139} & \textbf{0.232} & 0.181 & 0.266 & \textbf{0.179} & \textbf{0.263} & 0.165 & 0.266 & \textbf{0.156} & \textbf{0.252} & 0.453 & 0.484 & \textbf{0.436} & \textbf{0.481} \\
         & \multicolumn{1}{|c|}{192}& 0.160 & 0.248 & \textbf{0.157} & 0.250 & 0.183 & 0.272 & \textbf{0.182} & \textbf{0.268} & 0.183 & 0.283 & \textbf{0.177} & \textbf{0.270} & 0.382 & 0.430 & \textbf{0.379} & \textbf{0.424}  \\
         & \multicolumn{1}{|c|}{336}& 0.177 & 0.263 & \textbf{0.166} & \textbf{0.261} & 0.196 & 0.285 & \textbf{0.194} & \textbf{0.280} & 0.285 & 0.376 & \textbf{0.269} & \textbf{0.358} & 0.366 & 0.425 & \textbf{0.359} & \textbf{0.422}  \\
         & \multicolumn{1}{|c|}{720}& 0.208 & 0.290 & \textbf{0.187} & \textbf{0.281} & 0.234 & 0.314 & \textbf{0.232} & 0.312 & 0.285 & 0.376 & \textbf{0.269} & \textbf{0.358} & 0.366 & 0.425 & \textbf{0.359} & \textbf{0.422} \\
         \cmidrule(lr){2-18}
         & \multicolumn{1}{|c|}{Avg} & 0.173 & 0.260 & \textbf{0.162} & \textbf{0.256} & 0.199 & 0.284 & \textbf{0.197} & \textbf{281} & 0.211 & 0.308 & \textbf{0.200} & \textbf{0.293} & 0.430 & 0.468 & \textbf{0.382} & \textbf{437}  \\
        \specialrule{0.5pt}{1.5pt}{1.5pt}
        \multirow{5}{*}{\rotatebox{90}{Weather}} & \multicolumn{1}{|c|}{96} & 0.166 & 0.204 & \textbf{0.151} & \textbf{0.194} & 0.164 & 0.204 & \textbf{0.161} & \textbf{0.201} & 0.163 & 0.230 & \textbf{0.155} & \textbf{0.203} & 0.706 & 0.606 & \textbf{0.557} & \textbf{0.549}  \\
         & \multicolumn{1}{|c|}{192}& 0.208 & 0.242 & \textbf{0.196} & \textbf{0.238} & 0.205 & 0.239 & \textbf{0.202} & \textbf{0.234} & 0.223 & 0.294 & \textbf{0.214} & \textbf{0.258} & 1.221 & 0.893 & \textbf{0.739} & \textbf{0.666} \\
         & \multicolumn{1}{|c|}{336}& 0.257 & 0.280 & \textbf{0.248} & \textbf{0.275} & 0.256 & 0.279 & \textbf{0.251} & \textbf{0.278} & 0.258 & 0.320 & \textbf{0.250} & \textbf{0.280} & 0.884 & 0.732 & 0.916 & 0.765 \\
         & \multicolumn{1}{|c|}{720}& 0.316 & 0.322 & \textbf{0.311} & \textbf{0.319} & 0.316 & 0.320 & \textbf{0.315} & \textbf{0.319} & 0.325 & 0.376 & \textbf{0.320} & \textbf{0.330} & 0.892 & 0.748 & \textbf{0.584} & \textbf{0.587} \\
         \cmidrule(lr){2-18}
         & \multicolumn{1}{|c|}{Avg} & 0.237 & 0.262 & \textbf{0.226} & \textbf{0.257} & 0.235 & 0.261 & \textbf{0.232} & \textbf{0.258} & 0.242 & 0.305 & \textbf{0.235} & \textbf{0.268} & 0.926 & 0.745 & \textbf{0.699} & \textbf{0.642} \\
        \specialrule{0.5pt}{1.5pt}{1.5pt}
        \multirow{5}{*}{\rotatebox{90}{PEMS03}} & \multicolumn{1}{|c|}{12} & 0.065 & 0.171 & \textbf{0.059} & \textbf{0.164} & 0.088 & 0.206 & \textbf{0.086} & \textbf{0.206} & 0.083 & 0.194 & \textbf{0.075} & \textbf{0.182} & 0.108 & 0.207 & 0.109 & 0.211 \\
         & \multicolumn{1}{|c|}{24}& 0.093 & 0.206 & \textbf{0.076} & \textbf{0.185} & 0.144 & 0.263 & \textbf{0.140} & \textbf{0.261} & 0.117 & 0.238 & \textbf{0.102} & \textbf{0.217} & 0.126 & 0.230 & \textbf{0.124} & \textbf{0.228} \\
         & \multicolumn{1}{|c|}{48}& 0.161 & 0.274 & \textbf{0.123} & \textbf{0.234} & 0.275 & 0.367 & \textbf{0.262} & \textbf{0.357} & 0.205 & 0.319 & \textbf{0.156} & \textbf{0.271} & 0.164 & 0.265 & \textbf{0.160} & \textbf{0.261} \\
         & \multicolumn{1}{|c|}{96}& 0.348 & 0.456 & \textbf{0.208} & \textbf{0.303} & 0.508 & 0.528 & \textbf{0.467} & \textbf{0.501} & 0.259 & 0.363 & \textbf{0.213} & \textbf{0.326} & 0.189 & 0.283 & \textbf{0.184} & \textbf{0.279} \\
         \cmidrule(lr){2-18}
         & \multicolumn{1}{|c|}{Avg} & 0.167 & 0.277 & \textbf{0.116} & \textbf{0.221} & 0.254 & 0.341 & \textbf{0.239} & \textbf{0.331} & 0.166 & 0.278 & \textbf{0.136} & \textbf{0.249} & 0.147 & 0.246 & \textbf{0.144} & \textbf{0.245} \\
        \specialrule{0.75pt}{1.5pt}{1.5pt}
        \multicolumn{2}{c|}{Promotion} & --- & --- & \textbf{13.0\%} & \textbf{7.46\%} & --- & --- & \textbf{2.81\%} & \textbf{1.55\%} & --- & --- & \textbf{11.73\%} & \textbf{5.98\%} & --- & --- & \textbf{6.41\%} & \textbf{3.01\%} \\
        \bottomrule
    \end{tabular}
    }\end{center}
\end{table*}

We first focus on the evaluation of the ACE module. As summarized in Table \ref{table_ace}, integrating ACE leads to consistent improvements across a variety of Transformer-based models. These results demonstrate that ACE serves as a broadly applicable enhancement mechanism, capable of complementing different model designs without requiring structural modifications or hyperparameter tuning. Particularly for models such as vanilla Transformer and Crossformer, which primarily focus on leveraging strong attention mechanisms to capture and integrate cross-channel dependencies, there exists a relative weakness in modeling the intrinsic temporal structures of individual channels. As a result, these models often rely heavily on attention to compensate for insufficient input representations. By equipping them with the ACE module, which explicitly enriches the temporal expressiveness at the token level, we effectively enhance the quality of the input features before attention operations. This leads to more structured and informative interactions within the encoder, ultimately resulting in substantial performance gains across various datasets. Even for already strong baselines like iTransformer and PatchTST, ACE contributes consistent additional improvements, suggesting that the module generalizes well beyond its original setting and provides a versatile inductive bias for multivariate time series forecasting.

\begin{table}[H]
    \caption{
    Forecasting performance of iTransformer, PatchTST, and Transformer before and after integrating the Adaptive Channel Forecasting (ACF) module. Results are reported using MSE and MAE across four datasets: ETTh1, Weather, PEMS07, and Solar. To ensure fair comparison, we follow the same experimental settings as in our main results, without any additional hyperparameter tuning. The bottom row of ``Promotion" reports the relative improvements (\%) brought by ACF, calculated based on the original averaged MSE and MAE.
    }

    \label{table_acf}
    \begin{center}\resizebox{0.9\textwidth}{!}{
    \begin{tabular}{cc|cc>{\columncolor{shadow}}c>{\columncolor{shadow}}c|cc>{\columncolor{shadow}}c>{\columncolor{shadow}}c|cc>{\columncolor{shadow}}c>{\columncolor{shadow}}c}
        \toprule
         \multicolumn{2}{c|}{\multirow{2}{*}{Models}} & \multicolumn{2}{c}{\multirow{2}{*}{iTransformer}} & \multicolumn{2}{c|}{\textbf{w/ ACF}} & \multicolumn{2}{c}{\multirow{2}{*}{PatchTST}} & \multicolumn{2}{c|}{\textbf{w/ ACF}} & \multicolumn{2}{c}{\multirow{2}{*}{Transformer}} & \multicolumn{2}{c}{\textbf{w/ ACF}}\\
        & & \multicolumn{2}{c}{} & \multicolumn{2}{c|}{\textbf{(Ours)}} & \multicolumn{2}{c}{} & \multicolumn{2}{c|}{\textbf{(Ours)}} & \multicolumn{2}{c}{} & \multicolumn{2}{c}{\textbf{(Ours)}}\\
        \cmidrule(lr){3-6} \cmidrule(lr){7-10} \cmidrule(lr){11-14}
        \multicolumn{2}{c|}{Metric} & MSE & MAE & MSE & MAE & MSE & MAE & MSE & MAE & MSE & MAE & MSE & MAE  \\
        \specialrule{0.75pt}{0.0pt}{1.5pt}
        \multirow{5}{*}{\rotatebox{90}{ETTh1}} & \multicolumn{1}{|c|}{96} & 0.447 & 0.468 & \textbf{0.434} & \textbf{0.459} & 0.426 & 0.457 & \textbf{0.423} & \textbf{0.451} & 0.668 & 0.593 & \textbf{0.555} & \textbf{0.541 } \\
         & \multicolumn{1}{|c|}{192}& 0.495 & 0.503 & \textbf{0.478} & \textbf{0.494} & 0.466 & 0.489 & \textbf{0.462} & \textbf{0.486} & 0.788 & 0.646 & \textbf{0.723} & \textbf{0.615}  \\
         & \multicolumn{1}{|c|}{336}& 0.536 & 0.527 & \textbf{0.520} & 0.529 & 0.536 & 0.514 & \textbf{0.505} & \textbf{0.503} & 0.856 & 0.678 & \textbf{0.710} & \textbf{0.619}  \\
         & \multicolumn{1}{|c|}{720}& 0.665 & 0.603 & \textbf{0.630} & \textbf{0.602} & 0.659 & 0.612 & \textbf{0.654} & \textbf{0.608} & 0.946 & 0.738 & \textbf{0.833} & \textbf{0.682}  \\
         \cmidrule(lr){2-14}
         & \multicolumn{1}{|c|}{Avg} & 0.536 & 0.525 & \textbf{0.516} & \textbf{0.521} & 0.522 & 0.518 & \textbf{0.511} & \textbf{0.512} & 0.815 & 0.664 & \textbf{0.705} & \textbf{0.614}  \\
        
        \specialrule{0.5pt}{1.5pt}{1.5pt}
        \multirow{5}{*}{\rotatebox{90}{Weather}} & \multicolumn{1}{|c|}{96} & 0.166 & 0.204 & \textbf{0.153} & \textbf{0.199} & 0.164 & 0.204 & \textbf{0.150} & \textbf{0.198} & 0.706 & 0.606 & \textbf{0.230} & \textbf{0.297}  \\
         & \multicolumn{1}{|c|}{192}& 0.208 & 0.242 & \textbf{0.194} & \textbf{0.239} & 0.205 & 0.239 & \textbf{0.195} & 0.241 & 1.221 & 0.893 & \textbf{0.204} & \textbf{0.277}  \\
         & \multicolumn{1}{|c|}{336}& 0.257 & 0.280 & \textbf{0.248} & \textbf{0.277} & 0.256 & 0.279 & \textbf{0.239} & \textbf{0.275} & 0.884 & 0.732 & \textbf{0.265} & \textbf{0.328}  \\
         & \multicolumn{1}{|c|}{720}& 0.316 & 0.322 & \textbf{0.308} & \textbf{0.322} & 0.316 & 0.320 & \textbf{0.303} & 0.321 & 0.892 & 0.748 & \textbf{0.322} & \textbf{0.363}  \\
         \cmidrule(lr){2-14}
         & \multicolumn{1}{|c|}{Avg} & 0.237 & 0.262 & \textbf{0.226} & \textbf{0.259} & 0.235 & 0.261 & \textbf{0.222} & \textbf{0.259} & 0.926 & 0.745 & \textbf{0.255} & \textbf{0.316}  \\
        \specialrule{0.5pt}{1.5pt}{1.5pt}
        \multirow{5}{*}{\rotatebox{90}{Pems07}} & \multicolumn{1}{|c|}{12} & 0.067 & 0.166 & \textbf{0.060} & \textbf{0.147} & 0.091 & 0.209 & \textbf{0.070} & \textbf{0.168} & 0.166 & 0.229 & \textbf{0.087} & \textbf{0.190}  \\
         & \multicolumn{1}{|c|}{24}& 0.098 & 0.201 & \textbf{0.077} & \textbf{0.162} & 0.153 & 0.270 & \textbf{0.110} & \textbf{0.203} & 0.175 & 0.240 & \textbf{0.106} & \textbf{0.208}  \\
         & \multicolumn{1}{|c|}{48}& 0.179 & 0.292 & \textbf{0.164} & \textbf{0.288} & 0.303 & 0.381 & \textbf{0.188} & \textbf{0.253} & 0.181 & 0.248 & \textbf{0.142} & \textbf{0.243}  \\
         & \multicolumn{1}{|c|}{96}& 0.392 & 0.554 & \textbf{0.343} & \textbf{0.538} & 0.561 & 0.541 & \textbf{0.380} & \textbf{0.325} & 0.186 & 0.253 & \textbf{0.162} & 0.265  \\
         \cmidrule(lr){2-14}
         & \multicolumn{1}{|c|}{Avg} & 0.184 & 0.303 & \textbf{0.161} & \textbf{0.284} & 0.277 & 0.351 & \textbf{0.187} & \textbf{0.237} & 0.177 & 0.242 & \textbf{0.124} & \textbf{0.227}  \\
        \specialrule{0.5pt}{1.5pt}{1.5pt}
        \multirow{5}{*}{\rotatebox{90}{Solar}} & \multicolumn{1}{|c|}{96} & 0.211 & 0.247 & \textbf{0.203} & \textbf{0.237} & 0.215 & 0.276 & \textbf{0.207} & \textbf{0.258} & 0.735 & 0.769 & \textbf{0.683} & \textbf{0.643}  \\
         & \multicolumn{1}{|c|}{192}& 0.239 & 0.271 & \textbf{0.233} & \textbf{0.264} & 0.240 & 0.283 & 0.242 & 0.286 & 0.733 & 0.751 & \textbf{0.553} & \textbf{0.588}  \\
         & \multicolumn{1}{|c|}{336}& 0.258 & 0.285 & \textbf{0.249} & \textbf{0.276} & 0.263 & 0.300 & \textbf{0.250} & \textbf{0.293} & 0.757 & 0.778 & \textbf{0.737} & \textbf{0.752}  \\
         & \multicolumn{1}{|c|}{720}& 0.255 & 0.285 & \textbf{0.239} & \textbf{0.270} & 0.249 & 0.292 & \textbf{0.239} & \textbf{0.290} & 0.760 & 0.781 & \textbf{0.746} & \textbf{0.761}  \\
         \cmidrule(lr){2-14}
         & \multicolumn{1}{|c|}{Avg} & 0.241 & 0.272 & \textbf{0.231} & \textbf{0.262} & 0.242 & 0.288 & \textbf{0.235} & \textbf{0.282} & 0.746 & 0.770 & \textbf{0.680} & \textbf{0.686}  \\
        \specialrule{0.75pt}{1.5pt}{1.5pt}
        \multicolumn{2}{c|}{Promotion} & --- & --- & \textbf{4.24\%} & \textbf{1.86\%} & --- & --- & \textbf{3.51\%} & \textbf{1.34\%} & --- & --- & \textbf{31.6\%} & \textbf{25.3\%} \\
        \bottomrule
    \end{tabular}
    }\end{center}
\end{table}

Building upon the analysis of ACE, then turn to the evaluation of the ACF module. As shown in Table \ref{table_acf}, we observe that integrating the ACF module leads to consistent improvements. The performance gains are particularly notable when ACF is added to the vanilla Transformer, leading to a significant 21.8\% reduction in MSE and 15.3\% in MAE on average. This demonstrates that ACF is especially effective when applied to architectures that lack inherent mechanisms for modeling inter-channel dependencies. The benefit of ACF also extends to stronger baselines like PatchTST and iTransformer, where improvements, though smaller, remain consistent (e.g., 4.24\% MSE reduction for iTransformer). This suggests that ACF is not redundant but truly complementary: it refines the existing variable-wise mechanisms by learning a sparse, relevance-driven connectivity pattern rather than relying on uniform attention mixing. Moreover, ACF’s impact scales with dataset complexity. On PEMS07 (883 channels), ACF’s average 25.0\% MSE improvement underlines its power to suppress noise and filter out irrelevant channels via top-k selection, especially critical when so many variables could drown out the global signal. In periodic domains like Solar, where channel correlations ebb and flow over time, ACF dynamically adjusts which channels to attend to, capturing seasonality and trends while pruning away redundancy. Importantly, ACF achieves these gains with negligible additional overhead, and without requiring architectural changes to the attention layers, making it a versatile and lightweight module for multivariate time series forecasting.

\begin{figure}[H]
    \centering
    \includegraphics[width=1\linewidth]{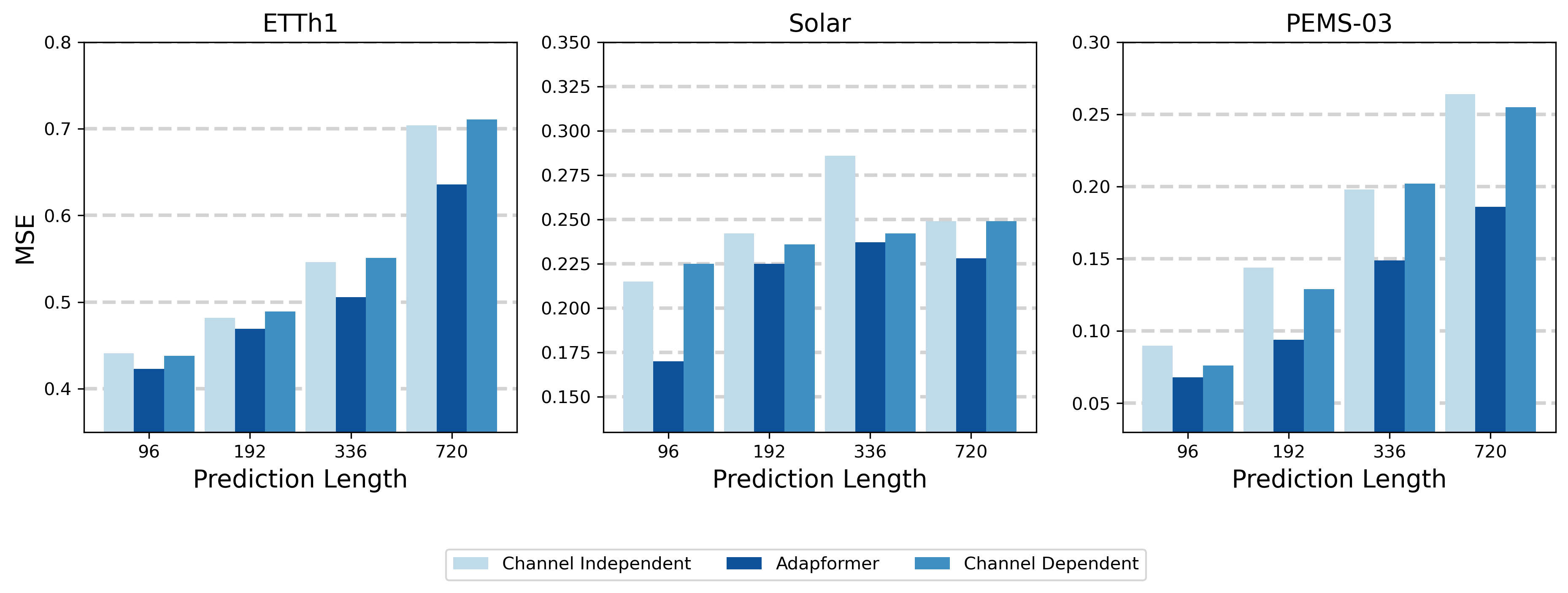}
    \caption{Comparison of Mean Square Errors (MSE) over varying prediction lengths on three benchmark datasets (ETTh1, Solar, and PEMS-03). We illustrate the performance of Channel Independent (\textbf{light blue}), Adapformer (\textbf{dark blue}), and Channel Dependent (\textbf{medium blue}). Adapformer consistently outperforms the two native channel strategies across all tested scenarios.}
    \label{fig16}
\end{figure}

Furthermore, our approach integrates both Channel-Independent and Channel-Dependent modeling strategies, enabling a direct comparison between the combined Adapformer and its pure CI or CD variants. This comparison is conducted by substituting the Adaptive Channel Forecaster module with traditional CI and CD predictors, respectively. Figure \ref{fig16} illustrates a bar chart where the dark blue bar in the center represents the MSE of our proposed Adapformer, while the bars on the left and right depict the MSEs of the CI and CD approaches. We evaluated the models on three datasets: ETTh1 (7 channels), Solar (137 channels), and PEMS-03 (358 channels). This selection of datasets ensures a comprehensive assessment of model performance across varying scales of channels. The results clearly demonstrate that our channel management strategy outperforms conventional naive approaches. This indicates that Adapformer is robust in handling diverse datasets, effectively maintaining channel-specific information in scenarios with a limited number of variables and mitigating cross-channel noise in datasets with a large number of channels. It is noteworthy that the performance gap on the ETTh1 dataset becomes more pronounced with longer prediction horizons. This underscores Adapformer's robust ability to capture and represent long-term temporal dependencies, a strength typically associated with CI modeling. Furthermore, on the PEMS-03 dataset, Adapformer consistently outperforms across all prediction lengths, with the CD approach only slightly surpassing the CI approach. This validates our analysis that, with a larger number of channels, modeling inter-variable dependencies is crucial for understanding latent patterns. Although the CD approach effectively leverages these dependencies, it still struggles with cross-variate noise. By addressing this critical challenge, Adapformer achieves superior performance compared to both conventional methodologies, demonstrating its efficacy in handling complex multivariate forecasting tasks.

\subsection{Varying Lookback Length}
\label{subsection_varying_lookback_length}

\begin{figure}
    \centering
    \caption{Forecasting performance with lookback lengths of $T=\{48,96,192,336,720\}$ with a fixed prediction horizon of 96 time steps. Results are compared among four Transformer-based models: Adapformer, iTransformer \cite{liu2023itransformer}, PatchTST \cite{nie2022time}, and CARD \cite{wang2024card}, across two datasets: ETTh1 and Electricity (ECL).}
    \includegraphics[width=0.78\linewidth]{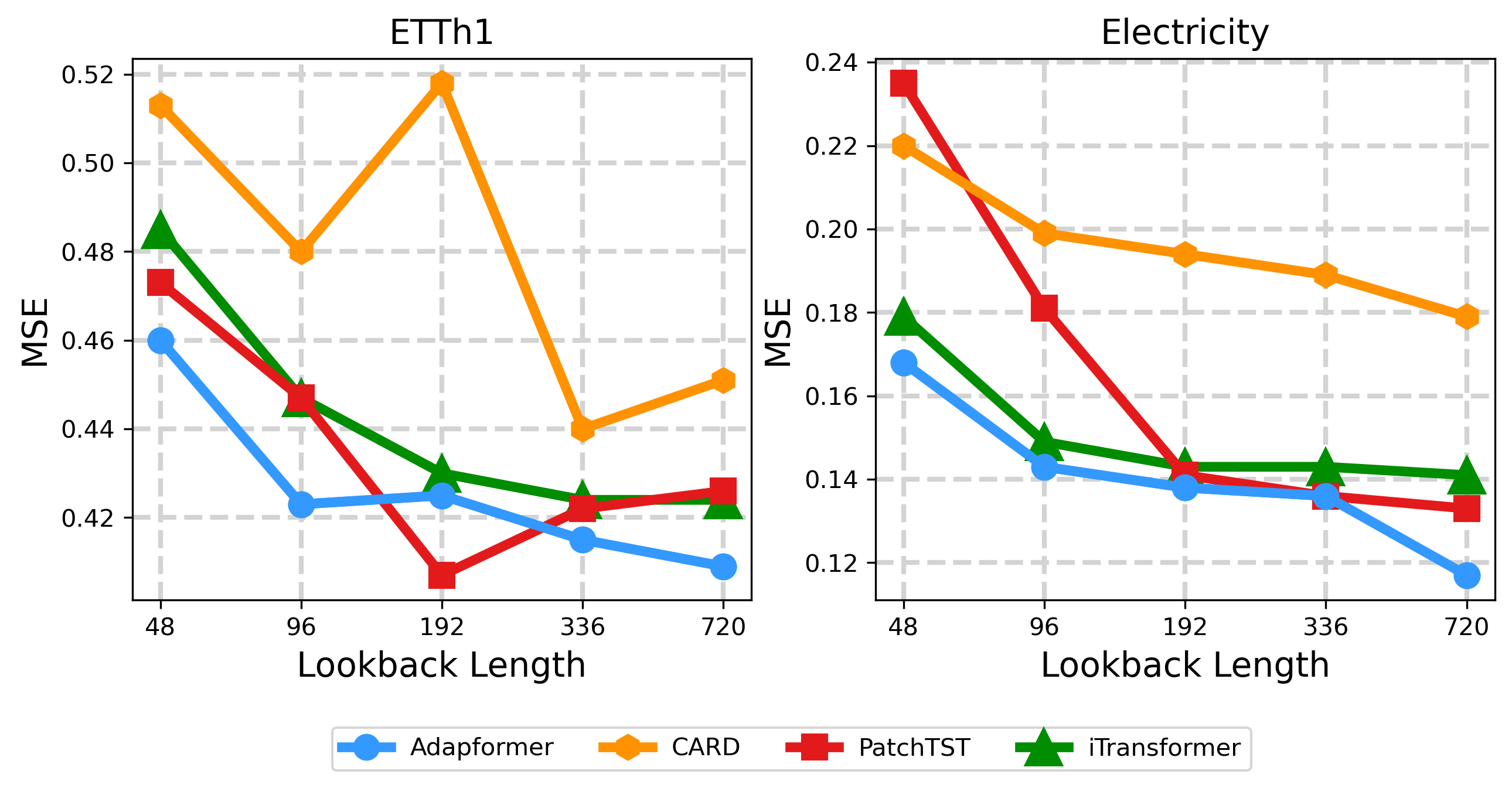}
    \label{fig17}
\end{figure}

Previous studies have demonstrated that increasing the lookback length in Transformer-based models does not consistently enhance forecasting performance \cite{nie2022time, zeng2023transformers}. This phenomenon is attributed to the escalation of complex dependencies and the inherent scalability limitations of existing architectures as the number of observations grows. To assess the robustness and scalability of our proposed Adapformer, we conducted an ablation study by systematically varying the lookback length—the temporal extent of input sequences provided to the model. Specifically, we tested lookback lengths of $T=\{48,96,192,336,720\}$ with a fixed prediction horizon of 96 time steps. Our evaluation encompassed four Transformer-based models: Adapformer, iTransformer \cite{liu2023itransformer}, PatchTST \cite{nie2022time}, and CARD \cite{wang2024card}, across two datasets: ETTh1 and Electricity (ECL).

As illustrated in Figure \ref{fig17}, Adapformer consistently outperforms the baseline models, exhibiting steadily lower Mean Squared Error (MSE) across all tested lookback lengths. This superior performance can be ascribed to Adapformer's sophisticated channel management strategy, which explicitly models stable multivariate correlations through the Similarity Block (SimBlock) and employs adaptive feature selection via the Adaptive Channel Feature (ACF) module. By effectively mitigating noise and preserving essential inter-variable relationships, Adapformer leverages extended temporal contexts to achieve more precise and reliable predictions. The consistent downward trend in MSE for Adapformer, even as lookback length increases, underscores the efficacy of our integrative approach. It is noteworthy that Adapformer's superiority becomes particularly pronounced with ultra-long input lengths. These findings highlight Adapformer's capability to maintain high forecasting performance amidst the intensified complexity of multivariate interactions, thereby validating the effectiveness of our channel management strategy in enhancing model scalability and robustness. Further experiments including hyperparameter sensitivity and model robustness can be found in \ref{appendix_hp_sensitivity} and \ref{appendix_robustness}.

\subsection{Complexity Analysis}

\begin{figure}[H]
    \centering
    \caption{Model Efficiency comparison across four datasets (ETTh1, ECL, Weather, and PEMS-07). For ETTh1, ECL, and Weather, each model uses an input sequence of length 96 and a forecasting horizon of 720 steps, while for PEMS-07 the prediction length is 96.}
    \includegraphics[width=\textwidth]{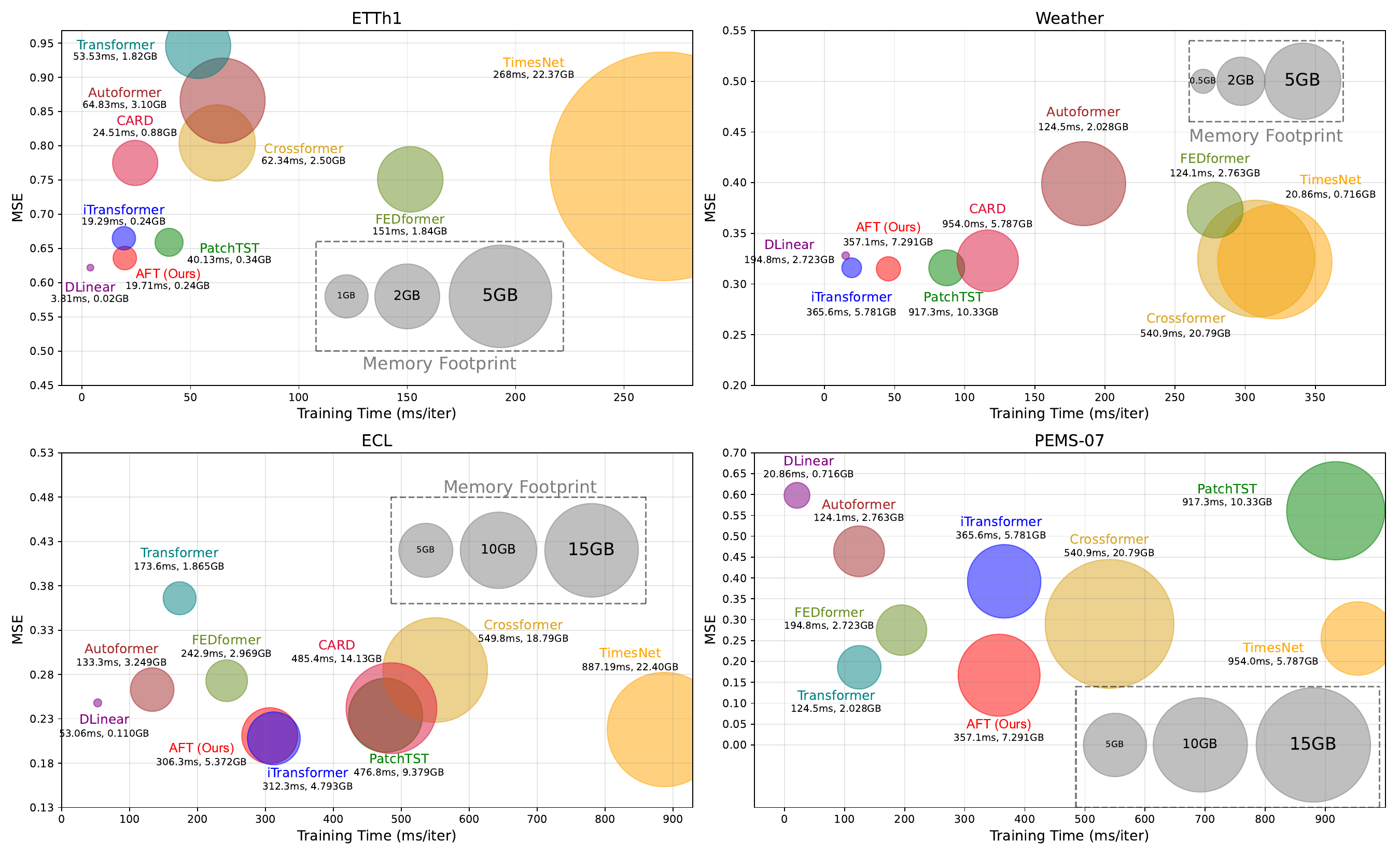}
    \label{complexity_comparison}
    \end{figure}

Adapformer does not alter the original \(\mathcal{O}(N^2)\) complexity of self-attention in the Transformer encoder, where $N$ represents the number of channels. Therefore we analytically present the computational complexity of the two targeted modules: ACE and ACF.
\begin{itemize}
    \item \textbf{ACE (Adaptive Channel Enhancer)} adopts a low-rank modeling strategy. Instead of constructing full-rank channel interactions, ACE projects channel embeddings into a lower-dimensional latent space of rank \(r\). This design intentionally avoids the quadratic interaction cost across channels, resulting in a time and space complexity of \(\mathcal{O}(rN)\), where \(r\) is a small constant relative to \(N\). Therefore, the enhancement step grows linearly with the number of channels.
    
    \item \textbf{ACF (Adaptive Channel Forecaster)} further optimizes the decoding process. Rather than aggregating information from all channels for forecasting each target series, ACF explicitly selects the top-\(k\) most relevant channels. This selection process reduces the forecasting computation to \(\mathcal{O}(kD)\) per channel, and \(\mathcal{O}(NkD)\) in total for all targets. As \(k\) remains a small fixed number compared to \(N\) and \(D\) is set as constant, the decoding complexity scales linearly with the input  dimensionality.
\end{itemize}

Thus, while the core Transformer retains its original attention complexity, the additional components ACE and ACF are specifically designed to introduce only lightweight, linear-in-\(N\) overheads. This architecture ensures that Adapformer can effectively scale to high-dimensional multivariate time series without incurring excessive computational or memory burdens.

In this experiment, we evaluated the complexity and efficiency of our model alongside the other baseline models across four diverse datasets: ETTh1, ECL, Weather, and PEMS-07. To ensure a fair and consistent comparison, all models were evaluated under uniform hyperparameters, including batch size of 32, forward dimension of 2048 and hidden dimension of 512. Furthermore, to fully assess the model's computational cost in scenarios involving large-scale data, we selected the longest prediction lengths for each dataset. Our evaluation specifically accounted for variations in dataset dimensionality; specifically, ETTh1 has 7 variables, Weather has 21 variables, ECL contains 321 variables and PEMS-07 includes as many as 883 variables. This broad range clearly illustrates our method's robustness and adaptability across datasets of varying complexities. The results demonstrate that our proposed model consistently achieves competitive forecasting accuracy  while maintaining significantly lower computational complexity compared to other advanced models, even in the highest-dimensional scenarios, as illustrated in Figure~\ref{complexity_comparison}. This highlights our model's capability to efficiently handle extensive multivariate time series data without sacrificing performance. In general, the consistent and robust performance observed across all experimental settings indicates that our model effectively balances prediction accuracy with computational efficiency, making it a promising candidate for real-world applications involving large-scale and diverse datasets.

\section{Conclusion and Future Work}
In this study, we introduced the Adapformer, a novel architecture designed to advance multivariate time series forecasting by effectively balancing channel-independent and channel-dependent strategies. Adapformer leverages a dual-stage encoder-decoder framework that selectively manages channel information, optimizing the utilization of relevant covariates while mitigating the impact of noise from excessive variables. This innovative selective channel management enhances the model's ability to capture intricate inter-variable dependencies and long-range temporal patterns, thereby improving forecasting accuracy and robustness. Empirical evaluations demonstrated that Adapformer consistently outperforms existing Transformer-based models, showcasing its superior scalability and efficiency in handling high-dimensional and noisy datasets across extended forecasting horizons. The model's ability to adaptively identify and prioritize the most pertinent covariates for each target variate enables it to maintain high predictive performance even as the complexity and dimensionality of the data increase. 

Looking forward, future work will explore several avenues to further enhance the capabilities of Adapformer. Firstly, optimizing the computational efficiency of the prediction phase through advanced parallelization techniques or approximation algorithms will be pursued to address scalability challenges associated with extremely large datasets. Secondly, the development of more sophisticated and robust similarity measurement methods, potentially incorporating causal analysis, will be investigated to improve the precision of covariate selection. Finally, integrating Adapformer with other advanced neural network architectures, such as graph neural networks, may provide further improvements in modeling complex inter-variable relationships, thereby broadening the model's applicability and effectiveness in diverse forecasting tasks. By addressing these areas, future research aims to refine Adapformer's performance and extend its utility, contributing to the ongoing advancement of multivariate time series forecasting methodologies.

\newpage\appendix

\section{Experimental Details}
\label{appendix_experimental_details}

\subsection{Datasets}
\label{appendix_datasets}
Here we provide the detailed introduction to the 5 widely-adopted benchmark datasets in MTSF that we use for our experiments:

\noindent\textbf{ETT}\footnote{\href{https://github.com/zhouhaoyi/ETDataset}{\texttt{https://github.com/zhouhaoyi/ETDataset}}}: The Electricity Transformer Temperature (ETT) dataset contains records of oil temperature and various power load conditions collected from two counties in China between July 2016 and July 2018. The dataset is divided into few subsets according to varying time granularities, from which we use two: ETTh1 and ETTh2, which are sampled hourly. We adopt this dataset from the Informer \cite{zhou2021informer}.

\noindent\textbf{ECL}\footnote{\href{https://archive.ics.uci.edu/ml/datasets/ElectricityLoadDiagrams20112014}{\texttt{https://archive.ics.uci.edu/ml/datasets/ElectricityLoadDiagrams20112014}}}: The Electricity Consuming Load (ECL) dataset captures the hourly electricity usage in kilowatt-hours (kWh) for 321 clients, spanning the period from 2012 to 2014. We adopt this dataset from the Informer \cite{zhou2021informer}.

\noindent\textbf{Weather}\footnote{\href{https://www.bgc-jena.mpg.de/wetter/}{\texttt{https://www.bgc-jena.mpg.de/wetter/}}}: The Weather dataset, provided by the Max Planck Biogeochemistry Institute in Germany, contains recordings of 21 meteorological variables, including air temperature and humidity. This climatological time series is sampled every 10 minutes over the full year of 2020. We adopt this dataset from the Informer \cite{zhou2021informer}.

\noindent\textbf{PeMS}:This transportation dataset, provided by the Caltrans Performance Measurement System (PeMS), captures traffic data from California, including metrics such as flow, occupancy, and speed. For our analysis, we utilize two public subsets: PEMS03 and PEMS07, both sampled at 5-minute intervals. We adopt this dataset from the SCINet \cite{liu2022scinet}.

\noindent\textbf{Solar Energy}\footnote{\href{http://www.nrel.gov/grid/solar-power-data.html}{\texttt{http://www.nrel.gov/grid/solar-power-data.html}}}: The Solar-Energy dataset records solar power production from 137 photovoltaic (PV) plants in Alabama State for the year 2006, with data sampled every 10 minutes. We adopt this dataset from the LSTNet \cite{lai2018modeling}.

\begin{table}[H]
    \captionsetup{width=\linewidth}
    \caption{Dataset Details}
    \label{table_datasets}
    \centering
    \resizebox{\textwidth}{!}
    {
    \begin{tabular}{l|c|c|c|c|c}
        \toprule
        Dataset & Dim & Input Length & Prediction Length & Dataset Size & Frequency  \\ 
        \toprule
        ETTh1, ETTh2 & $7$ & $96$ & $\{96, 192, 336, 720\}$ & $(12194, 2613, 2613)$ & Hourly \\
        \midrule
        Electricity & $321$ & $96$ & $\{96, 192, 336, 720\}$ & $(18412, 3945, 3947)$ & Hourly \\
        \midrule
        Weather & $21$ & $96$ & $\{96, 192, 336, 720\}$ & $(36887, 7904, 7905)$ & 10min \\
        \midrule
        PEMS-03 & $358$ & $96$ & $\{12, 24, 48, 96\}$ & $(18345, 3931, 3932)$ & 5min \\
        \midrule
        PEMS-07 & $883$ & $96$ & $\{12, 24, 48, 96\}$ & $(19756, 4233, 4235)$ & 5min \\
        \midrule
        Solar-Energy & $137$ & $96$ & $\{96, 192, 336, 720\}$ & $(36792, 7884, 7884)$ & 10min \\
        \bottomrule
    \end{tabular}
    }
\end{table}

\subsection{Implementation Details}
\label{appendix_implementation}

All the models and experimental frameworks are implemented entirely in Python 3.12.0 \cite{van1995python} and built upon PyTorch 2.4.0 \cite{paszke2019pyTorch}. All the experiments reported in this paper are conducted on a 16-core AMD EPYC 9654 CPU and a single NVIDIA RTX 4090 GPU. We select Adam \cite{kingma2017adam} as the optimizer with an initial learning rate in $\{5\times10^{-3},10^{-3},5\times10^{-4}\}$ and L2 loss combined with the auxiliary loss from Eq~\ref{aux_loss} to learn the model parameters. The learning rate is scheduled to follow an exponential decay pattern during training, which is halved at the end of each epoch. The number of training epochs is determined using an early stopping strategy, where the training is stopped when the model's performance (i.e. loss) ceases to improve on the validation set for a maximum of 3 times. The implementation of all baselines and their corresponding configurations are directly adopted from the \texttt{Time-Series-Library}\footnote{\href{https://github.com/thuml/Time-Series-Library.git}{\texttt{https://github.com/thuml/Time-Series-Library.git}}} provided by TimesNet \cite{wu2022timesnet}, which offers fair implementations of baseline methods based on the source code and configurations provided by each method's original paper.

\subsection{Hyperparameter Details}
\label{appendix_hyperparameter}

Table~\ref{table_hparam} lists all hyperparameters for Adapformer used in our experiments. The top two rows present the shared architectural hyperparameters applied across all datasets, including encoder and decoder layer counts (EncLayers \& DecLayers), dropout rate (Dropout), number of attention heads (NHeads), and feedforward dimension (FF Dim). These remain fixed throughout all experiments. The rest of the table contains training-specific hyperparameters, which are tuned individually based on different datasets and prediction length. These include the learning rate, model hidden dimension (Hidden Dim), batch size, and two key parameters introduced by our method: Rank, which controls the low-rank approximation dimension in the ACE module and governs intra-channel temporal capacity; and Top-k Channels, which determines the number of most correlated variables used in the ACF module.

\begin{table}[H]
    \centering
    \caption{Default and dataset-specific hyperparameter configurations used in our experiments. The top 2 row shows the default architectural settings shared across all datasets, while the rest section lists training-specific hyperparameters tuned with respect to each dataset and prediction length. ``Rank" denotes the low-rank approximation dimension $r$ used in the Adaptive Channel Enhancer (ACE) module. ``Top-k Channels" refers to the number of $k$ most correlated variables selected for each target by the Adaptive Channel Forecaster (ACF).}
    \label{table_hparam}
    \resizebox{0.9\textwidth}{!}
    {
    \small
    \begin{tabular}{l|c|c|c|c|c|c}
        \toprule
        \multicolumn{2}{c|}{H-params} & \makebox[6em][c]{\textbf{EncLayers}} & \makebox[6em][c]{\textbf{DecLayers}} & \makebox[6em][c]{\textbf{Dropout}} & \makebox[6em][c]{\textbf{NHeads}} & \makebox[6em][c]{\textbf{FF Dim}} \\
        \cmidrule{1-7}

        \multicolumn{2}{c|}{Default} & \makebox[6em][c]{2} & \makebox[6em][c]{1} & \makebox[6em][c]{0.1} & \makebox[6em][c]{8} & \makebox[6em][c]{2048} \\

        \toprule
        \multicolumn{2}{c|}{H-params} & \makebox[6em][c]{\textbf{Learning Rate}} & \makebox[6em][c]{\textbf{Hidden Dim}} & \makebox[6em][c]{\textbf{Batch Size}} & \makebox[6em][c]{\textbf{Rank}} & \makebox[6em][c]{\textbf{Top-k Channels}} \\
        \cmidrule{1-7}
        
        \multirow{4}{*}{\rotatebox{90}{ETTh1}}
        & 96  & $1\times 10^{-3}$ & 128 & 32 & 64 & 2 \\
        & 192  & $1\times 10^{-3}$ & 128 & 32 & 64 & 3 \\
        & 336  & $1\times 10^{-3}$ & 128 & 32 & 64 & 5 \\
        & 720  & $1\times 10^{-3}$ & 128 & 32 & 64 & 2 \\
        \midrule

       \multirow{4}{*}{\rotatebox{90}{ETTh2}}
        & 96  & $5\times 10^{-4}$ & 128 & 32 & 64 & 2 \\
        & 192  & $5\times 10^{-3}$ & 128 & 32 & 64 & 1 \\
        & 336  & $5\times 10^{-4}$ & 256 & 32 & 96 & 2 \\
        & 720  & $5\times 10^{-3}$ & 256 & 32 & 96 & 2 \\
        \midrule

        \multirow{4}{*}{\rotatebox{90}{ECL}}
        & 96  & $5\times 10^{-3}$ & 256 & 16 & 64 & 10 \\
        & 192  & $5\times 10^{-3}$ & 256 & 16 & 64 & 10 \\
        & 336  & $5\times 10^{-3}$ & 256 & 16 & 128 & 15 \\
        & 720  & $5\times 10^{-3}$ & 256 & 16 & 128 & 15 \\
        \midrule

        \multirow{4}{*}{\rotatebox{90}{Weather}}
        & 96  & $5\times 10^{-4}$ & 128 & 32 & 64 & 5 \\
        & 192  & $5\times 10^{-3}$ & 128 & 32 & 64 & 15 \\
        & 336  & $5\times 10^{-3}$ & 512 & 32 & 126 & 6 \\
        & 720  & $5\times 10^{-3}$ & 512 & 32 & 256 & 15 \\
        \midrule

        \multirow{4}{*}{\rotatebox{90}{PEMS-03}}
        & 12  & $1\times 10^{-3}$ & 512 & 16 & 64 & 8 \\
        & 24  & $1\times 10^{-3}$ & 512 & 16 & 64 & 8 \\
        & 48  & $1\times 10^{-3}$ & 512 & 16 & 128 & 8 \\
        & 96  & $1\times 10^{-3}$ & 512 & 16 & 128 & 8 \\
        \midrule

        \multirow{4}{*}{\rotatebox{90}{PEMS-07}}
        & 12  & $1\times 10^{-3}$ & 512 & 16 & 32 & 36 \\
        & 24  & $1\times 10^{-3}$ & 512 & 16 & 32 & 36 \\
        & 48  & $1\times 10^{-3}$ & 512 & 16 & 64 & 36 \\
        & 96  & $1\times 10^{-3}$ & 512 & 16 & 72 & 36 \\
        \midrule

        \multirow{4}{*}{\rotatebox{90}{Solar}}
        & 96  & $5\times 10^{-4}$ & 256 & 32 & 32 & 3 \\
        & 192  & $5\times 10^{-3}$ & 256 & 32 & 64 & 5 \\
        & 336  & $5\times 10^{-3}$ & 256 & 32 & 64 & 5 \\
        & 720  & $5\times 10^{-4}$ & 256 & 32 & 64 & 3 \\
        \bottomrule
    \end{tabular}
    }
\end{table}

\section{Hyperparameter Sensitivity}
\label{appendix_hp_sensitivity}
In this study, we concentrate on two pivotal hyperparameters: learning rate and model dimension. An optimal learning rate ensures efficient training, facilitating rapid convergence without overshooting minima. In our experiment, we manually selected four values of learning rate to tune: $\{1 \times 10^{-2}, 5 \times 10^{-3}, 1 \times 10^{-3}, 5 \times 10^{-4}\}$. The model dimension determines the capacity to capture and represent complex patterns within the data. We tune the model dimension in the range of $\{128, 256, 512, 1024\}$. We select four datasets: ETTh1, ETTh2, Weather and Solar for this test. Both the lookback length and the prediction length are set to $T = 96$.

\label{subsection_hyperparameter_sensitivity}
\begin{figure}[H]
    \centering
    \includegraphics[width=0.78\linewidth]{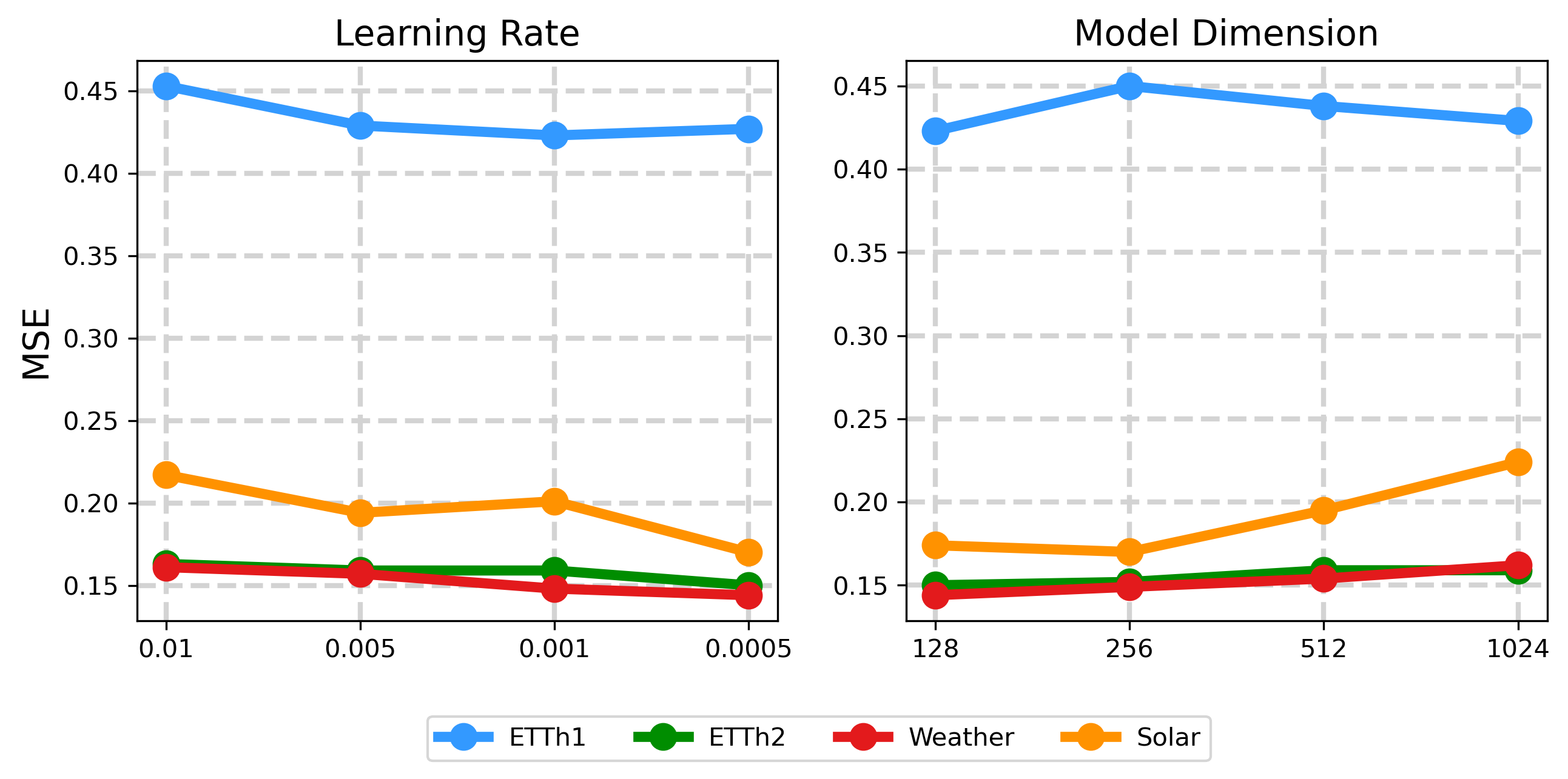}
    \caption[Hyperparameter Sensitivity]{Hyperparameter Sensitivity}
    \label{fig15}
\end{figure}

\begin{figure}[H]
    \centering
    \captionsetup{width=.9\linewidth}
    \caption{Sensitivity analysis of the top-$k$ selection parameter in ACF across three datasets (Weather, Solar, and ECL). All experiments are conducted with 96-input-96-output settings, with no additional hyperparameter tuned.}
    \includegraphics[width=0.9\textwidth]{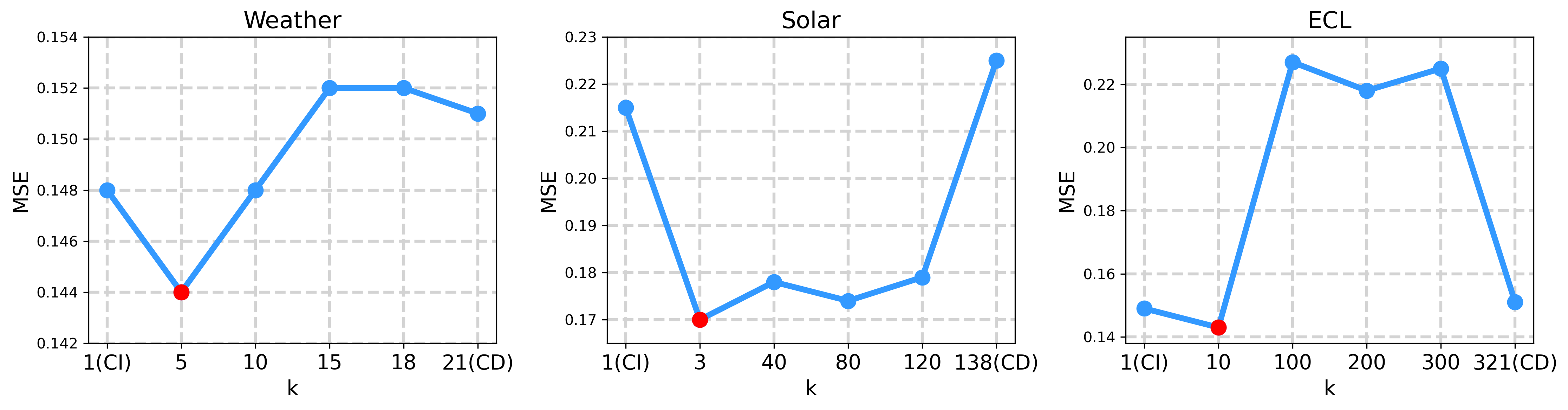}
    \label{complexity}
\end{figure}

We further investigate the sensitivity of the ACF module to the top-$k$ selection parameter as demonstrated in Figure~\ref{complexity}. Specifically, we vary $k$ from selecting only the target channel (Channel Independent) to selecting all channels (Channel Dependent), evaluating its impact on forecasting accuracy. As shown in Figure B.8, relatively low values of $k$ yield the best performance by balancing the trade-off between leveraging useful correlations and avoiding noise from irrelevant channels. This validates the effectiveness of ACF’s sparse relevance-driven connectivity design.

\begin{figure}[H]
    \centering
    \captionsetup{width=.9\linewidth}
    \caption{Sensitivity analysis of the low-rank dimension $r$ across five datasets. All experiments are conduct with 96 input length and 96 output length.}
    \includegraphics[width=0.6\textwidth]{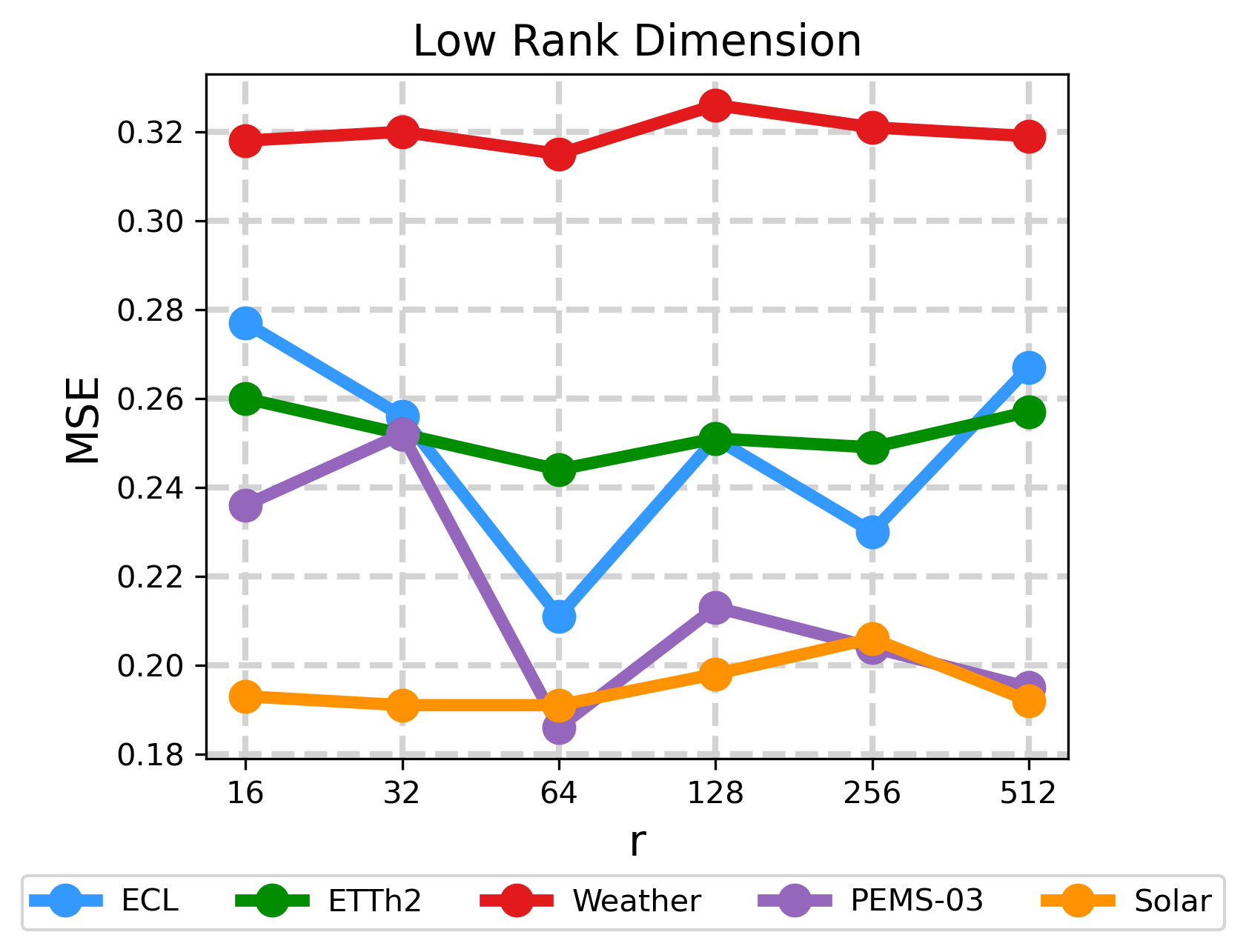}
    \label{r_complexity}
\end{figure}

Unlike general hyperparameter sensitivity where performance stability is desirable, here we aim to observe whether the low-rank dimension $r$ meaningfully influences performance, as it governs the expressiveness of ACE’s token enhancement. As shown in Figure~\ref{r_complexity}, variations in $r$ indeed lead to non-trivial changes in MSE across all datasets. For instance, both ECL and PEMS03 exhibit significant improvements when increasing $r$ from 32 to 64, but performance degrades again when $r$ becomes too large. This aligns with our theoretical intuition that $r$ controls the number of temporal variation modes injected into each token, and tuning it improperly may lead to underfitting (too small $r$) or overfitting/noise amplification (too large $r$). The observed fluctuations further verify that the low-rank projection dimension is an expressive and sensitive component, and thus should be carefully tuned based on data characteristics.

\section{Robustness Evaluation}
\label{appendix_robustness}
\begin{table}[H]
    \captionsetup{width=\linewidth}
    \caption{Adapformer's performance robustness. The results are obtained from 5 random runs.}
    \label{table_robustness}
    \centering
    \resizebox{\textwidth}{!}
    {
    \begin{tabular}{c|cc|cc|cc}
        \toprule
        Dataset & \multicolumn{2}{c}{ETTh1} & \multicolumn{2}{c}{ECL} & \multicolumn{2}{c}{Weather} \\
        \cmidrule(lr){2-3} \cmidrule(lr){4-5} \cmidrule(lr){6-7}
        Horizon & MSE & MAE & MSE & MAE & MSE & MAE \\
        \toprule
        96 & 0.423$\pm$ \colorbox{black!10}{0.001} & 0.442$\pm$ \colorbox{black!10}{0.002} & 0.143$\pm$ \colorbox{black!10}{0.000} & 0.242$\pm$ \colorbox{black!10}{0.000} & 0.144$\pm$ \colorbox{black!10}{0.002} & 0.195$\pm$ \colorbox{black!10}{0.001} \\
        192 & 0.469$\pm$ \colorbox{black!10}{0.002} & 0.485$\pm$ \colorbox{black!10}{0.002} & 0.159$\pm$ \colorbox{black!10}{0.002} & 0.251$\pm$ \colorbox{black!10}{0.000} & 0.208$\pm$ \colorbox{black!10}{0.003} & 0.260$\pm$ \colorbox{black!10}{0.002} \\
        336 & 0.506$\pm$ \colorbox{black!10}{0.003} & 0.527$\pm$ \colorbox{black!10}{0.002} & 0.174$\pm$ \colorbox{black!10}{0.001} & 0.262$\pm$ \colorbox{black!10}{0.001} & 0.256$\pm$ \colorbox{black!10}{0.004} & 0.273$\pm$ \colorbox{black!10}{0.001} \\
        720 & 0.636$\pm$ \colorbox{black!10}{0.003} & 0.611$\pm$ \colorbox{black!10}{0.003} & 0.211$\pm$ \colorbox{black!10}{0.003} & 0.281$\pm$ \colorbox{black!10}{0.004} & 0.315$\pm$ \colorbox{black!10}{0.001} & 0.320$\pm$ \colorbox{black!10}{0.002} \\
        \midrule
        Dataset & \multicolumn{2}{c}{PEMS03} & \multicolumn{2}{c}{PEMS07} & \multicolumn{2}{c}{Solar} \\
        \cmidrule(lr){2-3} \cmidrule(lr){4-5} \cmidrule(lr){6-7}
        Horizon & MSE & MAE & MSE & MAE & MSE & MAE \\
        \toprule
        12/96 & 0.068$\pm$ \colorbox{black!10}{0.002} & 0.171$\pm$ \colorbox{black!10}{0.001} & 0.065$\pm$ \colorbox{black!10}{0.001} & 0.164$\pm$ \colorbox{black!10}{0.001} & 0.170$\pm$ \colorbox{black!10}{0.001} & 0.235$\pm$ \colorbox{black!10}{0.002} \\
        24/192 & 0.094$\pm$ \colorbox{black!10}{0.002} & 0.204$\pm$ \colorbox{black!10}{0.002} & 0.089$\pm$ \colorbox{black!10}{0.001} & 0.193$\pm$ \colorbox{black!10}{0.003} & 0.225$\pm$ \colorbox{black!10}{0.002} & 0.272$\pm$ \colorbox{black!10}{0.001} \\
        48/336 & 0.149$\pm$ \colorbox{black!10}{0.002} & 0.259$\pm$ \colorbox{black!10}{0.002} & 0.127$\pm$ \colorbox{black!10}{0.002} & 0.234$\pm$ \colorbox{black!10}{0.002} & 0.237$\pm$ \colorbox{black!10}{0.002} & 0.276$\pm$ \colorbox{black!10}{0.001} \\
        96/720 & 0.186$\pm$ \colorbox{black!10}{0.004} & 0.301$\pm$ \colorbox{black!10}{0.003} & 0.167$\pm$ \colorbox{black!10}{0.001} & 0.270$\pm$ \colorbox{black!10}{0.001} & 0.191$\pm$ \colorbox{black!10}{0.002} & 0.252$\pm$ \colorbox{black!10}{0.001} \\
        \bottomrule
    \end{tabular}
    }
\end{table}

Table \ref{table_robustness} reports the error bars of Adapformer's forecasting performance, measured as the standard deviation over five random runs. The table shows that Adapformer maintains a high degree of stability across all datasets, with minimal variation in performance. These small standard deviations can be considered as random errors, well within an acceptable range relative to the original results.

\section{Showcases}
\label{appendix_showcase}

\begin{figure}[H]
    \centering
    \begin{minipage}[b]{0.3\textwidth}
        \centering
        \caption*{\textbf{Adapformer (Ours)}}
        \vspace{-0.25cm}
        \includegraphics[width=\textwidth]{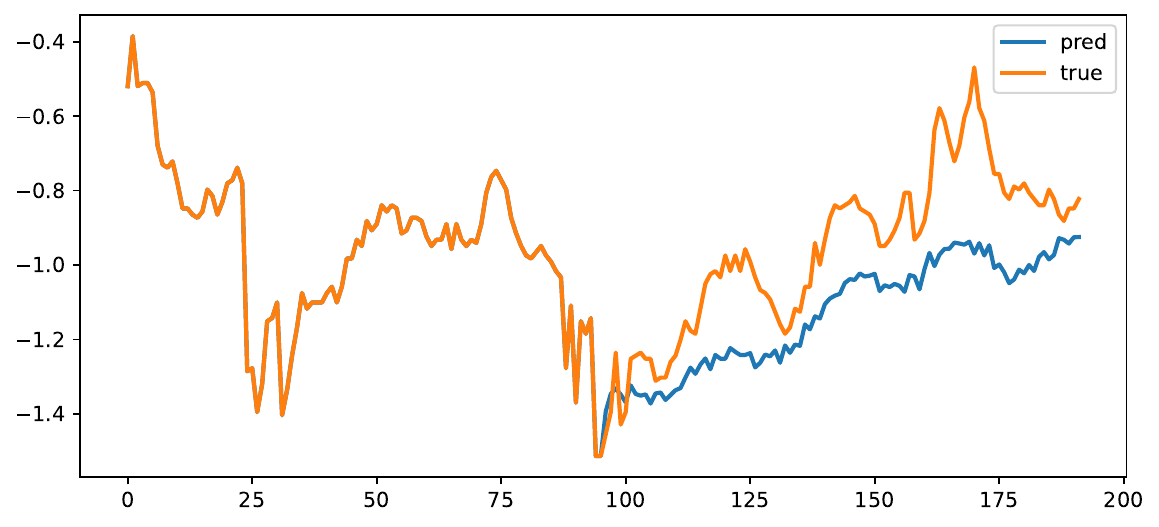}
    \end{minipage}
    \hfill
    \begin{minipage}[b]{0.3\textwidth}
        \centering
        \caption*{iTransformer}
        \vspace{-0.25cm}
        \includegraphics[width=\textwidth]{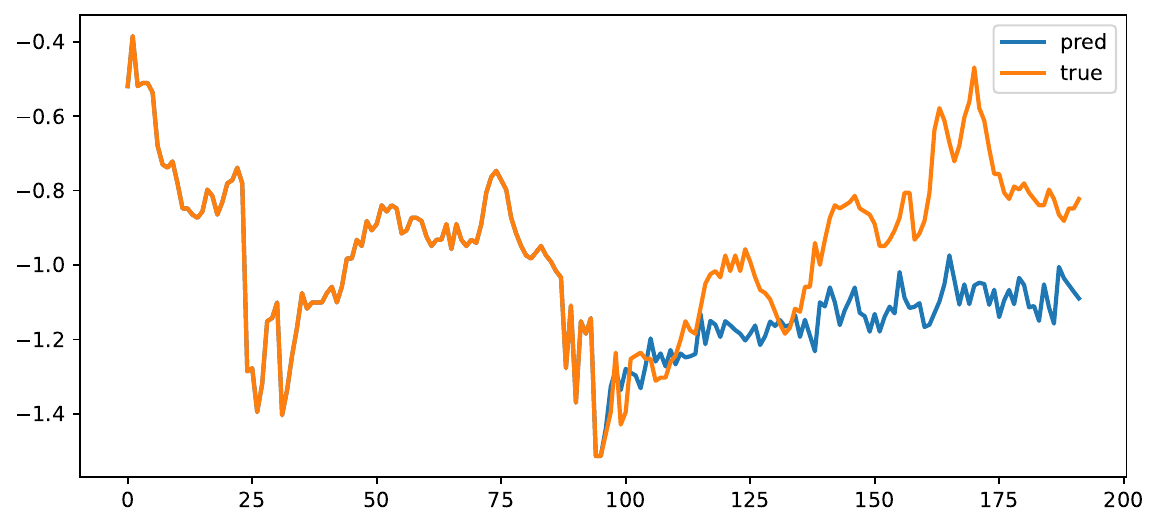}
    \end{minipage}
    \hfill
    \begin{minipage}[b]{0.3\textwidth}
        \centering
        \caption*{PatchTST}
        \vspace{-0.25cm}
        \includegraphics[width=\textwidth]{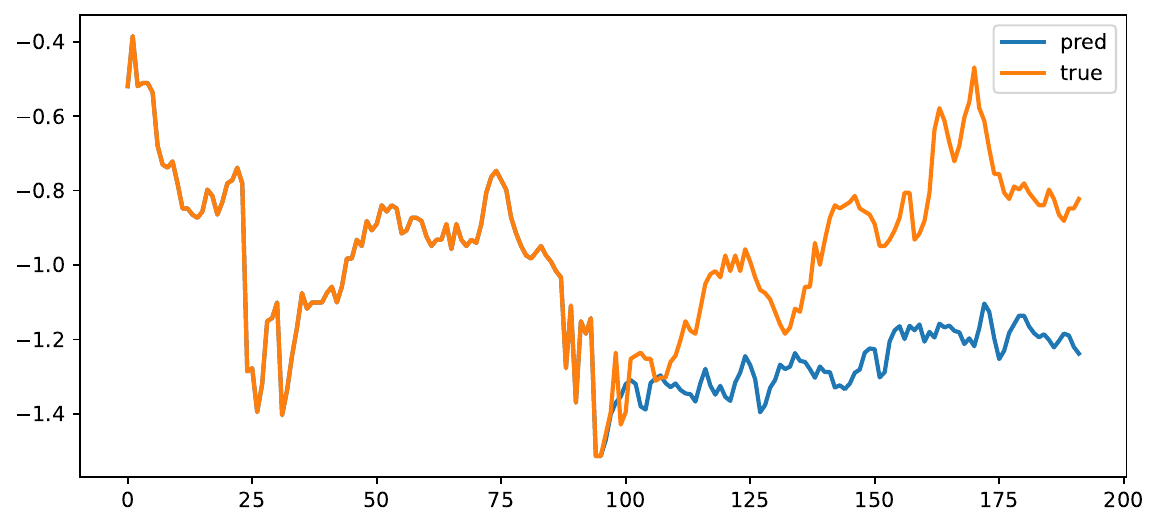}
    \end{minipage}
    \vfill
    \begin{minipage}[b]{0.3\textwidth}
        \centering
        \caption*{Crossformer}
        \vspace{-0.25cm}
        \includegraphics[width=\textwidth]{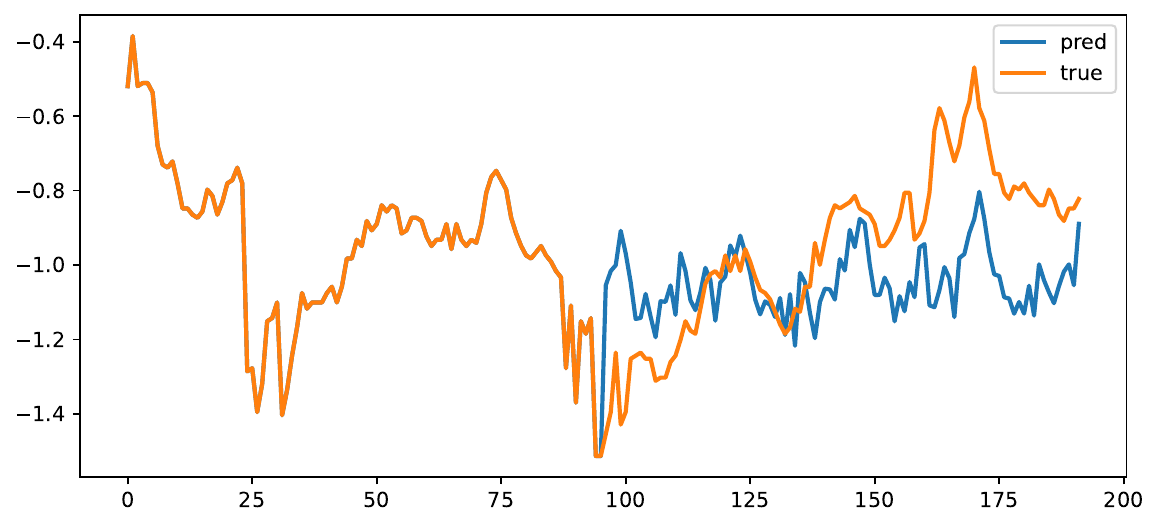}
    \end{minipage}
    \hfill
    \begin{minipage}[b]{0.3\textwidth}
        \centering
        \caption*{CARD}
        \vspace{-0.25cm}
        \includegraphics[width=\textwidth]{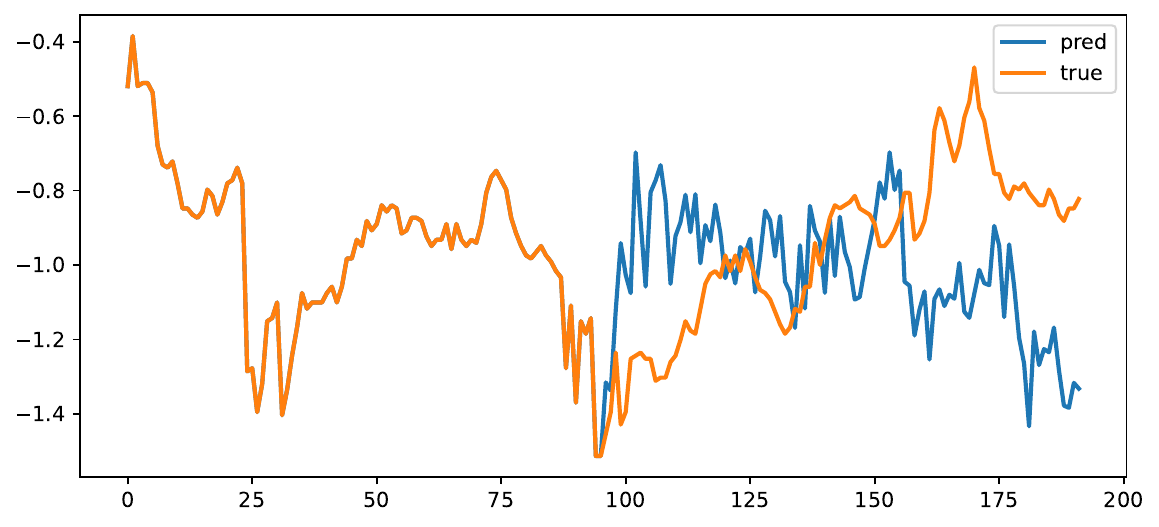}
    \end{minipage}
    \hfill
    \begin{minipage}[b]{0.3\textwidth}
        \centering
        \caption*{DLinear}
        \vspace{-0.25cm}
        \includegraphics[width=\textwidth]{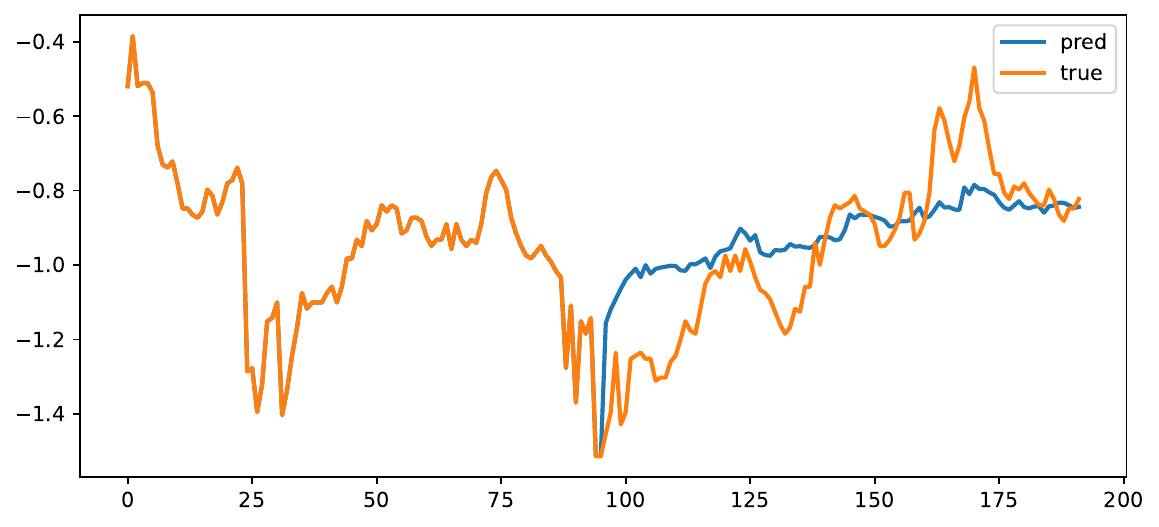}
    \end{minipage}
    \caption{Visualization of input-96-predict-96 results on the ETTh1 dataset}
    \label{etth1_showcase}
\end{figure}

\begin{figure}[H]
    \centering
    \begin{minipage}[b]{0.3\textwidth}
        \centering
        \caption*{\textbf{Adapformer (Ours)}}
        \vspace{-0.25cm}
        \includegraphics[width=\textwidth]{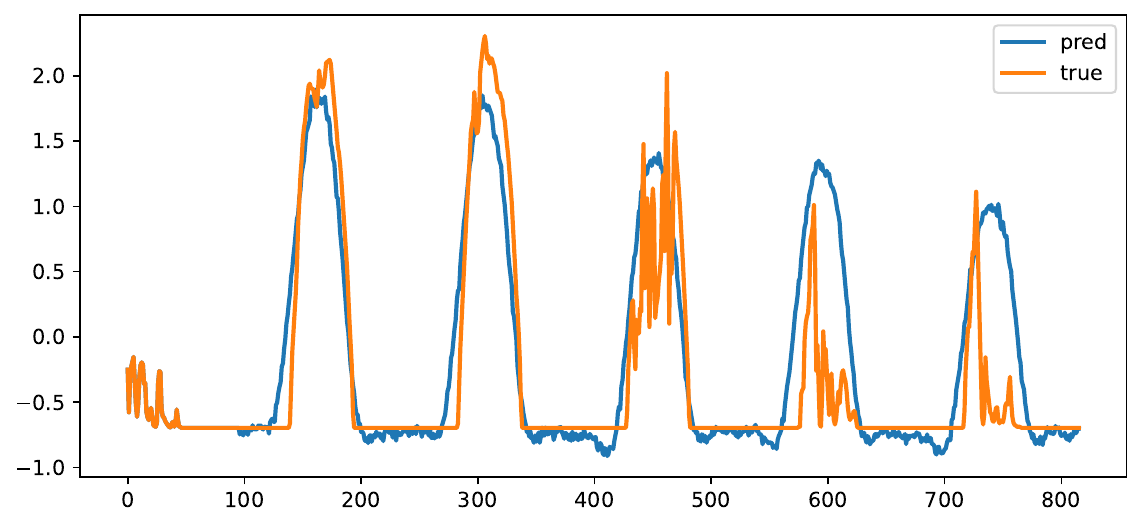}
    \end{minipage}
    \hfill
    \begin{minipage}[b]{0.3\textwidth}
        \centering
        \caption*{iTransformer}
        \vspace{-0.25cm}
        \includegraphics[width=\textwidth]{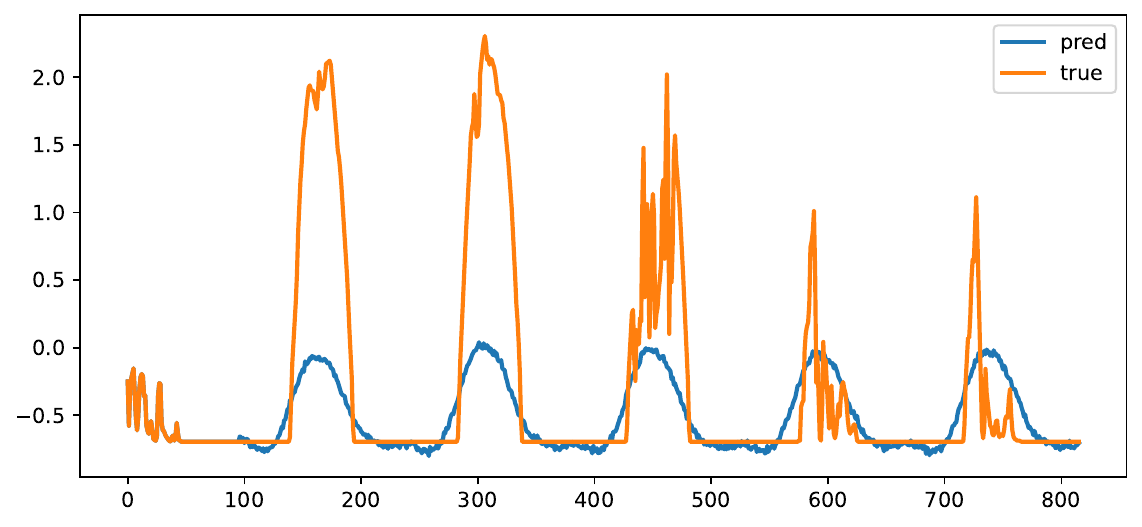}
    \end{minipage}
    \hfill
    \begin{minipage}[b]{0.3\textwidth}
        \centering
        \caption*{PatchTST}
        \vspace{-0.25cm}
        \includegraphics[width=\textwidth]{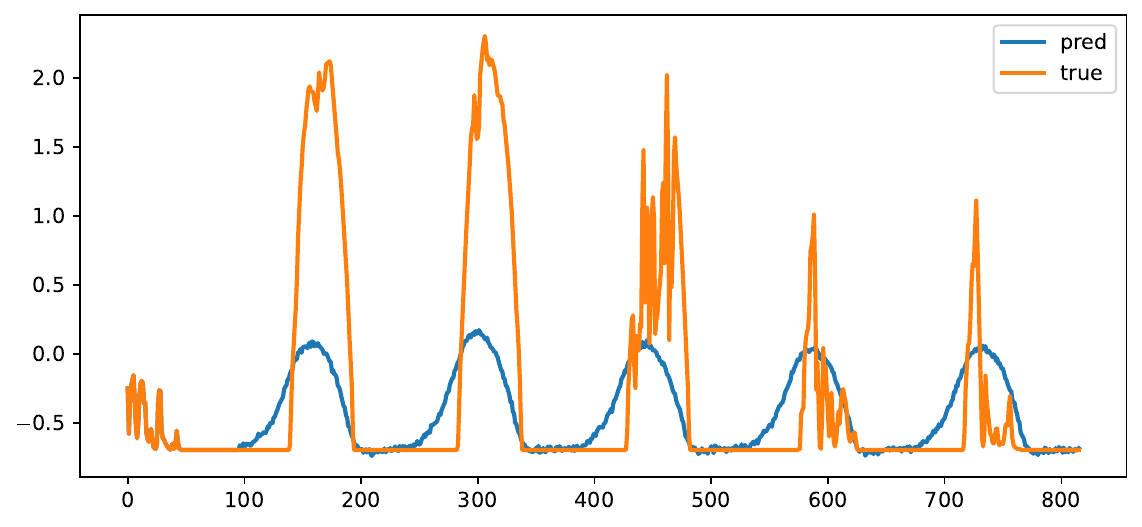}
    \end{minipage}
    \vfill
    \begin{minipage}[b]{0.3\textwidth}
        \centering
        \caption*{Crossformer}
        \vspace{-0.25cm}
        \includegraphics[width=\textwidth]{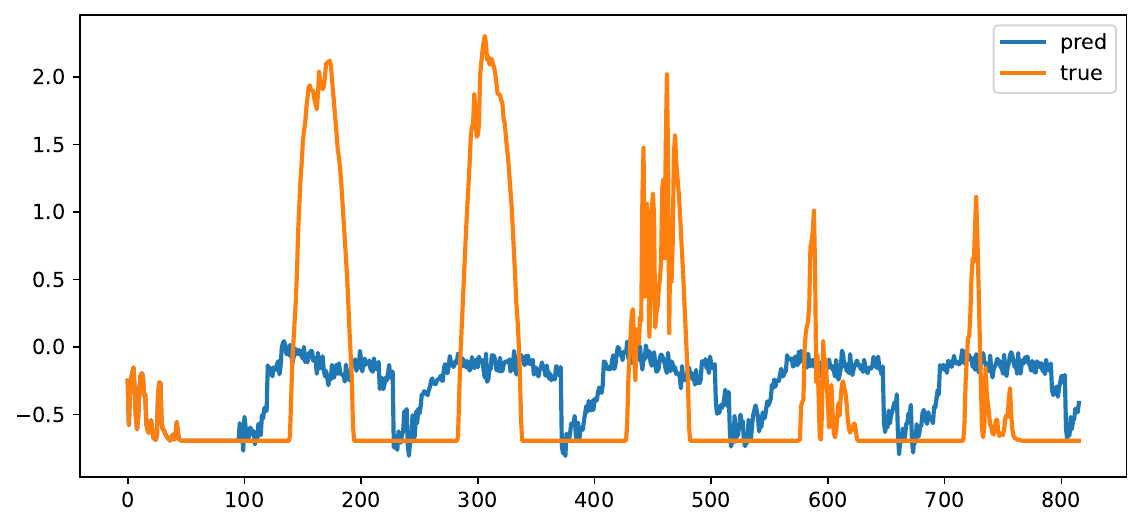}
    \end{minipage}
    \hfill
    \begin{minipage}[b]{0.3\textwidth}
        \centering
        \caption*{CARD}
        \vspace{-0.25cm}
        \includegraphics[width=\textwidth]{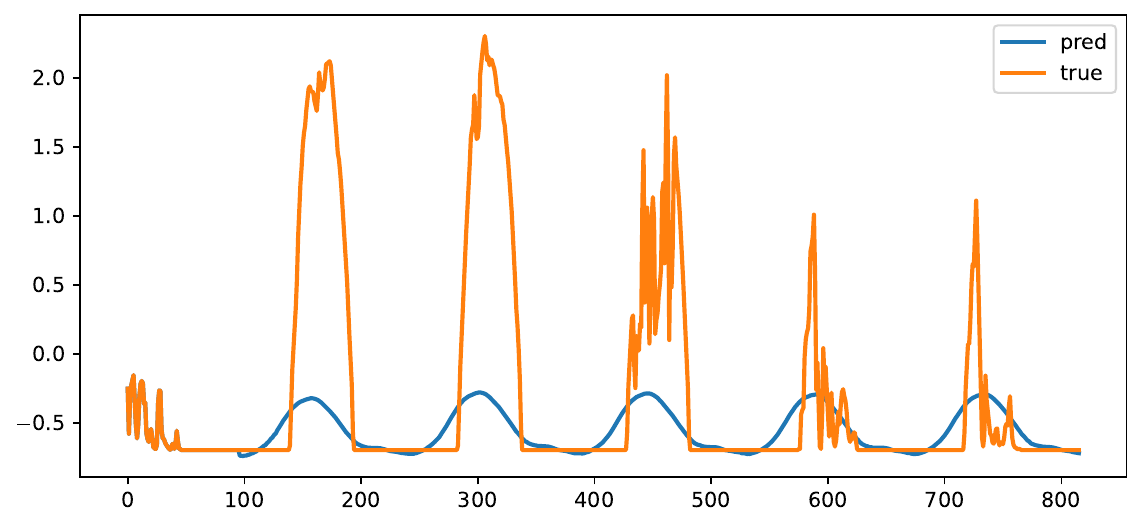}
    \end{minipage}
    \hfill
    \begin{minipage}[b]{0.3\textwidth}
        \centering
        \caption*{DLinear}
        \vspace{-0.25cm}
        \includegraphics[width=\textwidth]{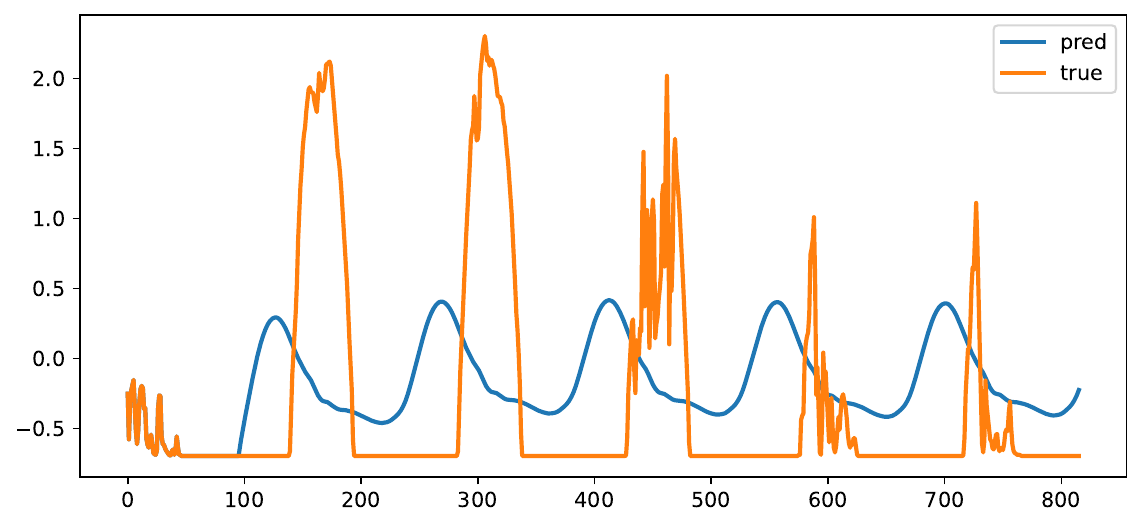}
    \end{minipage}
    \caption{Visualization of input-96-predict-720 results on the Solar dataset}
    \label{solar_showcase}
\end{figure}

We provide the prediction visualization examples of Adapformer, compared with some other state-of-the-arts models. Each plot shows the ground truth of a single target series in orange and our model prediction in blue. The left half of each plot corresponds to the input sequence, while the right half includes the predicted outputs. Clearly we can observe the accuracy and superiority of Adapformer. As shown in Figure~\ref{etth1_showcase} , in the case study of ETTh1, Adapformer more closely aligns with the overall trend of the ground truth. Particularly in the rising segments, our model captures key patterns of turning points with greater accuracy. Unlike PatchTST and iTransformer, which show under-responsive predictions to trend changes, or CARD, which suffers from excessive noise, Adapformer produces stable yet responsive forecasts. Compared to the overly smoothed outputs of DLinear, its predictions maintain a better balance between smoothness and fidelity to the true dynamics. On the Solar dataset with long-horizon forecasting, as illustrated in Figure~\ref{solar_showcase}, Adapformer provides significantly more faithful and stable predictions on the fluctuations compared to other models. It captures both the periodic structure and the amplitude of the signals with high accuracy, while maintaining low noise during the flat, inter-peak intervals. While most baseline models are able to roughly recover the periodicity, they consistently underestimate the signal amplitude. iTransformer and PatchTST perform relatively well in terms of maintaining stability in low-activity regions, but fail to reach the correct peak heights. In contrast, Crossformer suffers from severe fluctuations due to noisy outputs, and DLinear generates predictions that are systematically phase-shifted, failing to align with the true signal structure. These results highlight Adapformer’s ability to generalize well over long forecasting horizons in highly periodic scenarios.

\newpage
\section{Algorithm Pseudocode}
\label{appendix_pseudocode}

\begin{figure}[H]
    \centering
    \includegraphics[width=1\linewidth]{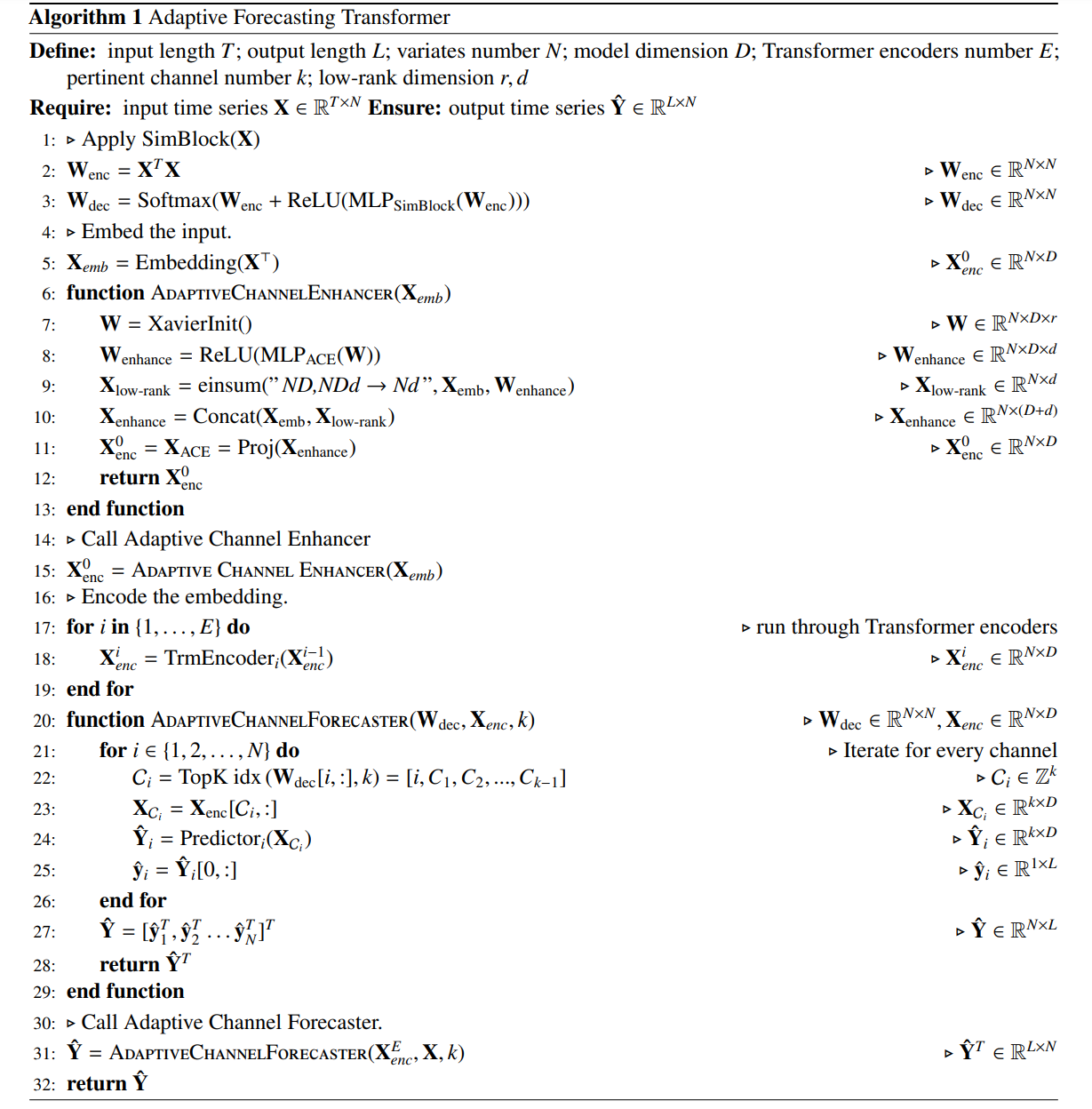}
    \label{pseudocode}
\end{figure}

\newpage
\bibliographystyle{unsrt}  
\bibliography{main} 

\end{document}